\documentclass[10pt,twocolumn,letterpaper]{article}

\usepackage[pagenumbers]{cvpr} %
\usepackage{wrapfig}
\usepackage[utf8]{inputenc} %
\usepackage[T1]{fontenc}    %
\usepackage{url}            %
\usepackage{booktabs}       %
\usepackage{amsfonts}       %
\usepackage{amssymb}
\usepackage{amsmath}
\usepackage{nicefrac}       %
\usepackage{microtype}      %
\usepackage[dvipsnames]{xcolor}
\usepackage{graphicx}
\usepackage{multirow}
\usepackage{subcaption}
\usepackage{longtable}
\usepackage{duckuments}
\usepackage{soul}
\usepackage{listings}
\usepackage{colortbl}

\definecolor{codegreen}{rgb}{0,0.6,0}
\definecolor{codegray}{rgb}{0.5,0.5,0.5}
\definecolor{codepurple}{rgb}{0.58,0,0.82}
\definecolor{backcolour}{rgb}{0.95,0.95,0.92}

\lstdefinestyle{mystyle}{
    backgroundcolor=\color{backcolour},   
    commentstyle=\color{codegreen},
    keywordstyle=\color{magenta},
    numberstyle=\tiny\color{codegray},
    stringstyle=\color{codepurple},
    basicstyle=\ttfamily\footnotesize,
    breakatwhitespace=false,         
    breaklines=true,                 
    captionpos=b,                    
    keepspaces=true,                 
    numbers=left,                    
    numbersep=5pt,                  
    showspaces=false,                
    showstringspaces=false,
    showtabs=false,                  
    tabsize=1
}

\newcommand{\e}[1]{#1}
\newcommand{\ee}[1]{#1}

\newcommand{\eee}[1]{#1}

\newcommand{\ev}[1]{#1}

\definecolor{cvprblue}{rgb}{0.21,0.49,0.74}
\usepackage[pagebackref,breaklinks,colorlinks,allcolors=cvprblue]{hyperref}

\def\paperID{14457} %
\def\confName{CVPR}
\def\confYear{2026}

\title{If you can describe it, they can see it: Cross-Modal Learning of Visual Concepts from Textual Descriptions}

\author{Carlo Alberto Barbano\\
University of Turin \\
\and
Luca Molinaro\\
University of Turin \\
\and
Massimiliano Ciranni\\
University of Genoa \\
\and
Emanuele Aiello\\
Politecnico di Torino \\
\and
Vito Paolo Pastore\\
University of Genoa \\
\and
Marco Grangetto\\
University of Turin \\
}

\begin{document}
\maketitle
\begin{abstract}
Humans can visualize new and unknown concepts from their natural language description, based on their experience and previous knowledge. Insipired by this, we present a way to extend this ability to Vision-Language Models (VLMs), teaching them novel concepts by only using a textual description. We refer to this approach as Knowledge Transfer (KT). Our hypothesis is that the knowledge of a pre-trained VLM can be re-used to represent previously unknown concepts. Provided with a textual description of the novel concept, KT works by aligning relevant features of the visual encoder, obtained through model inversion, to its text representation.
Differently from approaches relying on visual examples or external generative models, KT transfers knowledge within the same VLM by injecting visual knowledge directly from the text.
Through an extensive evaluation on several VLM tasks, including classification, segmentation, image-text retrieval, and captioning, we show that: 1) KT can efficiently introduce new visual concepts from a single textual description; 2) the same principle can be used to refine the representation of existing concepts; and 3) KT significantly improves the performance of zero-shot VLMs.
\end{abstract}

\section{Introduction}
\label{sec:intro}

\begin{figure*}
    \centering
    \begin{subfigure}[t]{0.48\textwidth}
        \centering
        \includegraphics[width=\linewidth,trim=20 0 19 0,clip]{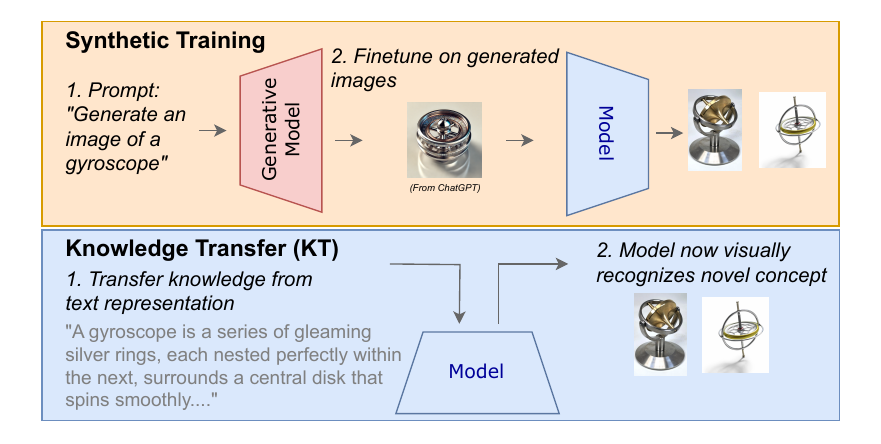}
        \caption{\textbf{Our research question:} \emph{Can we leverage existing knowledge in a pre-trained VLM to teach it novel concepts?} Differently from training with synthetic data or generative models~\citep{chen2023unified, hammoud2024synthclip}, with Knowledge Transfer we aim at leveraging existing knowledge within a model to teach it novel concepts, or improve existing ones.}
        \label{fig:kt-vs-synth}
    \end{subfigure}
    \hfill
    \begin{subfigure}[t]{0.48\textwidth}
        \centering
        \includegraphics[width=\linewidth]{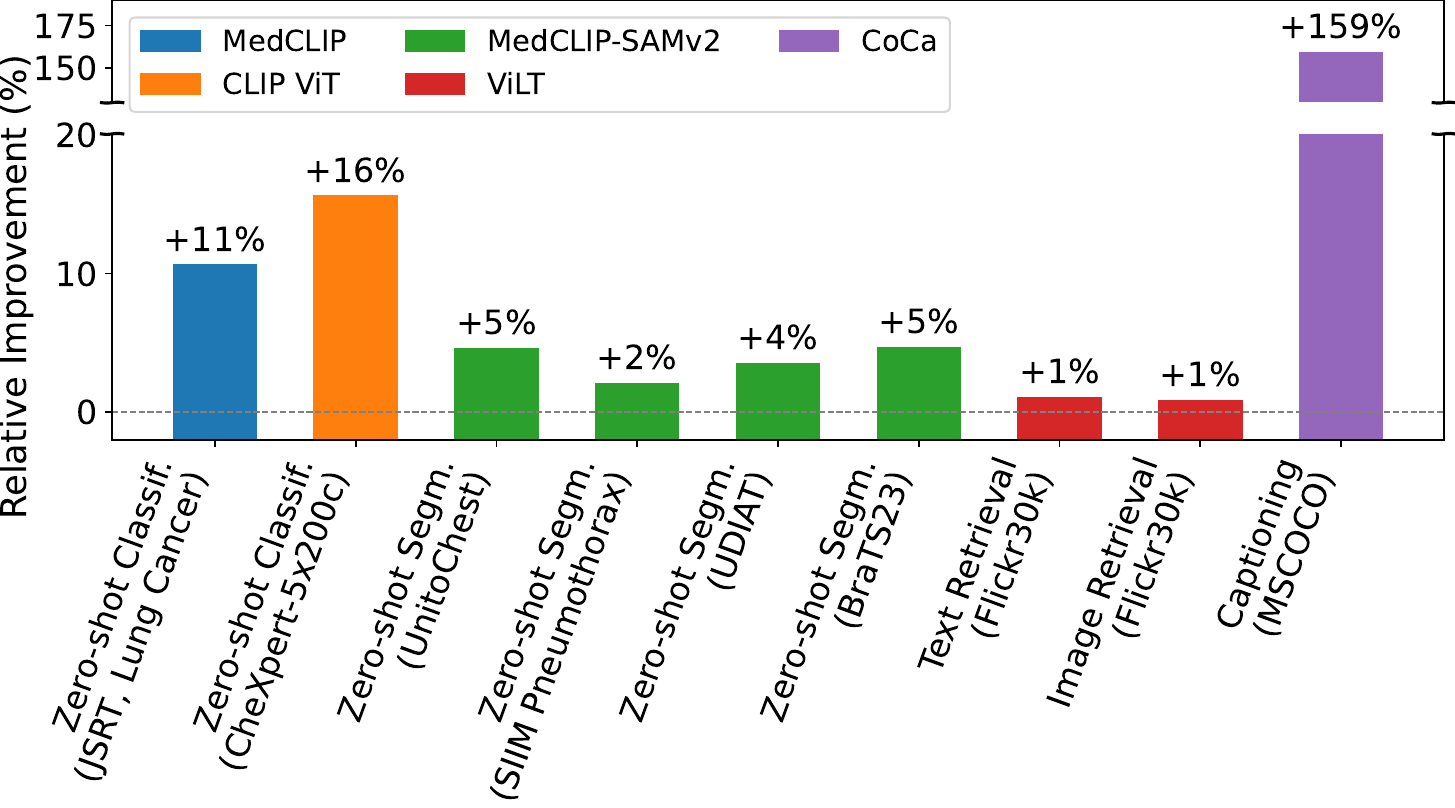}
        \caption{\textbf{Knowledge Transfer improves performance across a variety of tasks and architectures.} Differently from existing approaches, such as K-LITE~\citep{shen2022k}, LLaMP~\citep{zheng2024large}, and SynthClip~\citep{hammoud2024synthclip}, this is achieved without using any real image nor generative model.}
        \label{fig:results-overview}
    \end{subfigure}
    \caption{\textbf{Overview of Knowledge Transfer (KT):} \subref{fig:kt-vs-synth} shows our research question, while \subref{fig:results-overview} summarizes performance improvements.}
    \label{fig:overview}
\end{figure*}

Can a blind person who gained sight recognize the objects they previously knew only by touch? This is a philosophical riddle posed by William Molyneux in 1668 to John Locke~\citep{locke1948essay}, which has been relevant in vision neuroscience for decades. Recent research has shown that, while this does not happen immediately after sight restoration, cross-modal mappings develop rapidly in human subjects, within days~\citep{held2011newly}. While recent research in multimodal neural networks has focused on this cross-modal interaction~\citep{schwettmann2023multimodal}, in this paper we aim to answer a slightly revisited version of Molyneux's riddle, in which our model already has some previous knowledge of the world. \e{We hypothesize that prior knowledge of a pre-trained VLM can be used to produce a reasonable visual representation of an unknown concept,} if an explicative textual description is provided. This prior knowledge can be obtained with multimodal pre-training, for example by employing image-text alignment as done in CLIP and other similar works~\citep{radford2021learning, girdhar2023imagebind, yu2022coca, kim2021vilt}.
Learning novel concepts starting from a textual description has been explored in different works such as SynthCLIP~\citep{hammoud2024synthclip} and others~\citep{sandfort2019data, zhang2020medical, jahanian2022generative,zhou2023training, chen2023unified}. These approaches, however, rely on the availability of generative models, which is not trivial in low-data contexts such as medical imaging. 
\e{Furthermore, instead of just retrieving concepts learned during pre-training as in zero-shot VLMs~\cite{radford2021learning}, we explicitly incorporate new visual concepts by injecting novel textual descriptions into the VLM, without relying on external visual data or generative model (Fig.~\ref{fig:kt-vs-synth}). This enables performance improvements across a wide range of downstream tasks without using any real images, as shown in Fig.~\ref{fig:results-overview}. To the best of our knowledge, this is the first work to exploit existing knowledge of a VLM to learn novel concepts across modalities.} %

\begin{figure}
    \centering
    \begin{subfigure}[t]{0.47\linewidth}
        \includegraphics[width=\linewidth]{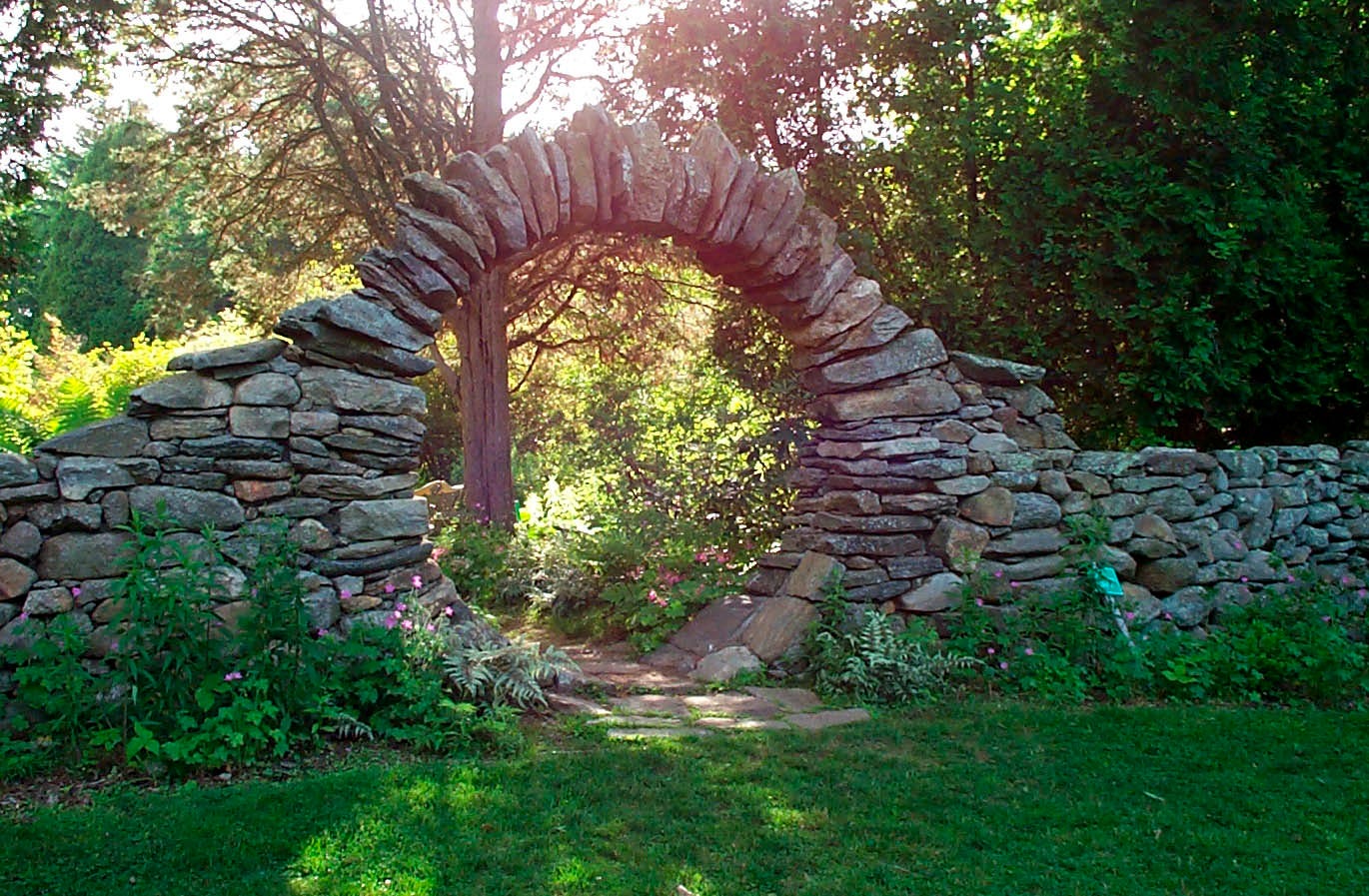}
        \caption{\textbf{CLIP (B)} Top-3 zero-shot predictions: \emph{Triumphal Arch, Stone Wall, Steel Arch Bridge}.\\
        \textbf{CLIP (B) + KT} Top-3 zero-shot predictions: \emph{\underline{Moongate}, Triumphal Arch, Stone Wall}}
    \end{subfigure}
    \hfill
    \begin{subfigure}[t]{0.47\linewidth}
        \includegraphics[width=\linewidth]{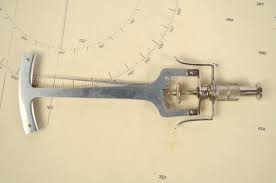}
        \caption{\textbf{CLIP (L)} Top-3 zero-shot predictions: \emph{(Cocktail Shaker, Odometer, Dragonfly}.\\
        \textbf{CLIP (L) + KT} Top-3 zero-shot predictions: \emph{\underline{Tonometer}, Cocktail Shaker, Espresso Maker}}
    \end{subfigure}
    \caption{\textbf{Knowledge Transfer can introduce novel concepts in a multimodal model}, by leveraging prior visual knowledge of the visual encoder and a textual description of the target concept. %
    In the example, a CLIP model~\citep{radford2021learning} learns the concepts \emph{Moongate} and \emph{Tonometer}, without using any real image, while retaining a good accuracy on general zero-shot classification (58.10\% vs 56.43\% and 70.79\% vs 70.61\% on ImageNet-1k).}
    \label{fig:teaser}
\end{figure}

Leveraging natural language supervision to learn novel visual concepts is a process we call \emph{Knowledge Transfer} (KT), \e{inspired by previous research on zero-shot modality generalization~\cite{elhoseiny2013write}}. 
An illustrative example of the goal of KT is shown in Fig.~\ref{fig:teaser}, where a CLIP-based zero-shot classifier is presented with unknown concepts. %
In this work, we propose a novel framework for KT that does not require parameters to be shared across visual and textual encoders, thus being general with respect to the VLM architecture. Specifically, starting from the textual description of the novel concepts, we synthesize matching imaging via model inversion~\citep{kazemi2024we}, later employed to fine-tune the model with a visual-text matching loss. 
\\Our findings show that KT:
\begin{enumerate}
    \item Successfully introduces novel concepts in pre-trained VLMs with only textual descriptions;
    \item Improves the visual accuracy on already existing concepts;
    \item Improves zero-shot downstream tasks such as classification, segmentation, and image-text retrieval and shows potential for out-of-domain generalization.
\end{enumerate}

\section{Related Works}
\label{sec:related}

Multimodal representation learning aims to bridge the gap between different modalities (e.g. visual and textual), enabling models to process them jointly. VLMs like CLIP~\citep{radford2021learning}, CoCa~\citep{yu2022coca}, Flamingo~\citep{alayrac2022flamingo} and ImageBind~\citep{girdhar2023imagebind}, align visual and textual features in a shared embedding space, empowering zero-shot and few-shot learning in various visual tasks.
\eee{Efforts have been made to understand how VLMs internally process cross-modal information (e.g. multimodal neurons)~\citep{goh2021multimodal, schwettmann2023multimodal, pan2023finding}.} \eee{Here, we provide an overview of works related to KT in different fields.}

\paragraph{Cross-Modal Transfer}
Cross-modal knowledge distillation~\citep{gupta2016cross, tang2021vidlankd, huo2024c2kd} is a strategy for transferring knowledge between modalities, to enrich representations. Methods like VidLanKD~\citep{tang2021vidlankd} and C2KD~\citep{huo2024c2kd} employ modality-bridging techniques to improve generalization in zero and few-shot scenarios. These approaches typically require substantial multimodal data and complex training procedures. In contrast, our method uses textual descriptions to introduce new visual concepts with minimal data by efficient reuse of prior knowledge. %
Lin~\emph{et al.}~\citep{lin2023multimodality} recently showed that integrating cues from multiple modalities can enhance concept learning, mirroring human learning.
Their approach leverages few-shot examples of paired multimodal data to enhance unimodal downstream tasks. Differently from them, we use single-modal text data to introduce new visual knowledge.

\paragraph{Text-based \e{zero-shot} methods} Methods such as K-LITE~\citep{shen2022k} leverage structured text rather than simple captions as in CLIP~\citep{radford2021learning}, to improve pre-training. With a similar goal, Zheng~\emph{et al.}~\citep{zheng2024large}, propose to augment zero-shot prompts with an LLM, or to jointly fine-tune it in their LLaMP framework, to obtain richer classification prompts. Differently from these approaches, we use text data to describe novel concepts without using any real image.
Additionally, text-only methods have been proposed to \e{achieve} visual understanding. For example, CapDec~\citep{nukrai2022text} \e{and CLOSE~\cite{gu2023can}}, leverage the alignment between the visual and text encoder in CLIP \e{by training a downstream model (e.g. for captioning or VQA) on text embedding and then using it on images.} \e{Our approach, on the other hand, aims at achieving knowledge transfer in the VLM itself via fine-tuning.} %

\paragraph{Synthetic Training} Other works \citep{chen2023unified, zhou2023training, jahanian2022generative} involve synthetic data generation to train discriminative models. For example, SynthCLIP~\citep{hammoud2024synthclip} leverages Stable Diffusion \citep{rombach2021high} to train a CLIP model entirely on synthetic data. While effective, this approach depends on the quality and diversity of generated data. Our approach differs by integrating novel concepts into existing models without relying on external knowledge of a text-to-image generative model and a computationally expensive data generation pipeline.

\paragraph{Incremental Learning} 
\ee{Due to the similar objective, which is introducing novel concepts, we could also frame KT as a form of incremental learning. Recent works such as CLIP~\citep{liu2025cclip}, TPPT~\citep{lu2025tppt}, and ENGINE~\citep{zhou2025engine} exploit textual prompts or external knowledge to stabilize updates across tasks.
Other approaches incorporate language cues or generative models to enrich class representations~\citep{cao2024gmm,zhang2025textualpriors,dalessandro2023multimodal}.} %
\e{Despite these advances, there is still a notable gap: none (to our knowledge) explicitly considers incremental introduction of new visual classes solely via textual descriptions, with no visual exemplars, and then deploy the model to recognise images of the new class (i.e., a “text-only → visual class incremental” paradigm). Our method fills precisely this gap, by using textual descriptions of novel concepts (without real images) to integrate them into a pre-trained VLM, thereby enabling zero- or few-shot recognition of the new class while avoiding large-scale retraining or storage of exemplar images.}

\begin{figure*}
    \includegraphics[width=\textwidth]{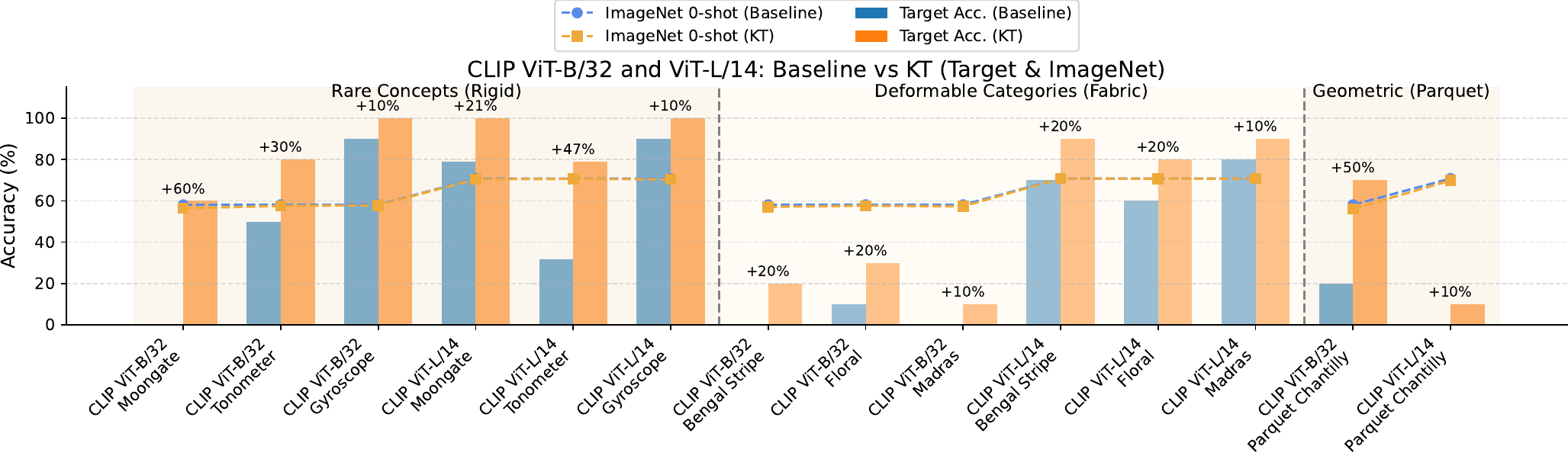}
    \caption{\eee{\textbf{Knowledge Transfer (KT) on novel and rare concepts} (high-level and fine-grained concepts). KT achieves improvements (even notable) in the target accuracy on the novel concept in all instances. We also make sure that catastrophic forgetting does not occur by monitoring zero-shot accuracy on ImageNet, which remains comparable with the baseline.}}
    \label{fig:kt-results-rare-concepts-all}
\end{figure*}

\section{Knowledge Transfer}

In this section, we present our proposed method for Knowledge Transfer, which we will simply refer to as KT.

\subsection{Goal of KT}
\label{sec:method}

Let $f_T: \mathbb{R}^L \rightarrow \mathbb{R}^n$ be a text encoder (where $L$ is the sequence length) and $f_V: \mathbb{R}^{w\times h} \rightarrow \mathbb{R}^n$ be a visual encoder (with $w$ and $h$ being the size of the image), our goal is to introduce new concepts into $f_V$ through $f_T$ using only the text modality. Let $X_T$ be a set of unpaired captions pertaining to a novel concept that we want to learn, and $X_V^*$ a set of \emph{ideal} ground truth images corresponding to that concept. What we would like to achieve is:
\begin{equation}
\begin{split}    
    \underbrace{sim(f_v(x_v^*), f_t(x_t))}_{s_t} - \underbrace{sim(f_v(x_v^*), f_t(x_k))}_{s_k} > 0 \\
    \forall x_v^* \in X_V^*,\, x_t \in X_T,\, x_k \in X_K 
    \label{eq:concept-alignment}
\end{split}
\end{equation}
\noindent where $X_K$ is the set of all other captions pertaining to other concepts ($X_K\,\cap\,X_T = \varnothing$). The condition in Eq.~\ref{eq:concept-alignment} means that all ideal visual samples should be mapped closer to the true corresponding captions then all other captions. In practice, if $X^*_V$ is available, we can satisfy Eq.~\ref{eq:concept-alignment} by optimizing its approximation~\citep{barbano2023unbiased}:
\begin{equation}
    \min_{f_V} -\frac{1}{|X^*_V|}\sum_{x^*_v}\frac{1}{|X_T|}\sum_{x_t}\log\frac{\exp(s_t)}{\exp(s_t) + \sum_{x_k}\exp(s_k)} \\
    \label{eq:clip-loss}
\end{equation}
\noindent which corresponds to the InfoNCE loss~\citep{radford2021learning, chen2020simple, khosla2020supervised}. 
In our setting, however, $X^*_V$ is not available, thus we propose to estimate it (e.g. with model inversion~\citep{kazemi2024we}) and then use the estimated values to jointly train $f_V$ and $f_T$ with the contrastive approach using InfoNCE~\citep{radford2021learning, girdhar2023imagebind}. 

As a practical example, a caption $x_t$ for the concept \emph{Moongate} could be \emph{``A perfectly circular archway built from uniformly cut stones or bricks, set into a larger wall. It forms a smooth circle, framing views of gardens or landscapes beyond, creating a picturesque portal.''}. More examples can be found in the supplementary material. %
\ee{As shown by this example, this method requires the visual encoder to have some prior knowledge about the visual concepts contained in the caption (e.g., stones, walls, circles and gardens). %
We argue that this is reasonable, as humans also struggle to visualize unfamiliar concepts, especially in specialized domains beyond their prior experience. Nonetheless, in the results section we provide an experiment on out-of-domain KT, showing how the proposed approach can still reach zero-shot out-of-domain generalization.}

\subsubsection{Estimating $X^*_v$ by inversion}
\label{sec:method-inversion}
The most straightforward way to estimate $X^*_v$ is to compute it by inverting the visual encoder $f_V$ starting from the textual embeddings of $X_T$, in order to obtain an approximation $\hat{X}^*_V \approx X^*_V$. To do so, we solve the following optimization problem, starting from random noise: 
\begin{equation}
\begin{split}
    \hat{X}^*_V = f_V^{-1}(f_T(X_T)) \qquad\qquad\qquad\qquad\qquad\qquad \\
    \approx \max_{\hat{X}^*_V}sim(A(f_V(\hat{X}^*_V)), f_T(X_T)) + \alpha R(\hat{X}^*_V)
    \label{eq:inversion}
\end{split}
\end{equation}
\noindent where, as in~\citep{kazemi2024we}, $f^{-1}$ is the inversion operator, $A$ is a random augmentation operation applied at each step  (e.g. random affine) and $R$ is a regularization term based on Total Variation (TV)~\citep{mordvintsev2015inceptionism}, weighted by $\alpha$.
Augmentation and regularization help in producing more naturally-looking images. 

\paragraph{Inversion as an effective alternative to generative models} 
Learning new visual concepts from natural language description can be solved, in principle, by synthesizing training images based on textual descriptions, employing generative models (e.g. DALL-E~\citep{ramesh2021zero} or Stable Diffusion~\citep{podell2023sdxl}). As such, a question that may automatically arise is why not employing such generative models for KT. First, using external generative models for augmenting the training set~\citep{sandfort2019data, zhang2020medical, jahanian2022generative,zhou2023training, chen2023unified, hammoud2024synthclip} is fundamentally different from the posed research question, as shown in Fig.~\ref{fig:kt-vs-synth}, which is transferring the knowledge from textual to visual modality within the same model, thus being independent from the availability of any external models. Besides, there are at least two other concerncs on the direct usage of generative models for the task at hand: \emph{i.}) we do not know if they already include the target concepts in their training set; \emph{ii.)} even if the target concepts are available in the original training set, in settings such as low-data domains (e.g. medical imaging), the availability of a generative model is not trivial, especially for clinically-relevant conditions. On top of this, employing external generative models would require more efforts in terms of computational time and resources.   %

\subsubsection{Finetuning on the new concepts}
\label{sec:method-finetuning}
After images $\hat{X}^*_V$ have been synthesized via model inversion, we can use them to train $f_v$ and $f_T$ with an image-text alignment objective such as InfoNCE (Eq.~\ref{eq:clip-loss}). In order to successfully match visual features to the desired concepts, we augment $X_T$ by prepending the corresponding concept name to each caption. In the example presented earlier, the fine-tuning caption would be represented by \emph{``A moongate is a perfectly circular archway built from uniformly cut stones [...]''}. This step is required in order to finally map the already learned low-level visual concepts to the high-level one itself.

\section{Experimental Setup}
\label{sec:setup}

We use several datasets to benchmark KT: an original dataset
\emph{RareConcepts} (for proof-of-concept experiments), ImageNet-1k~\citep{deng2009imagenet}, CheXpert-2x500c~\citep{huang2021gloria}, JSRT~\citep{shiraishi2000development}, UnitoChest~\citep{chaudhry2022unitochest}, UDIAT~\citep{yap2017automated}, SIIM Pneumothorax~\citep{zawacki2019siim}, BraTS23 Glioma~\citep{adewole2023brain}, Flickr30k~\citep{young-etal-2014-image}, and MSCOCO~\citep{lin2014microsoft}. We target many different tasks, such as classification, segmentation, captioning and text-image retrieval, across both natural and medical images. A more detailed presentation of the datasets can be found in the supplementary material, along with a detailed description of the training details. In summary, we produce the target concepts' captions using a mix of LLM-based text and handcrafted captions. Inversion is run for 5k steps, and a quick fine-tuning of the visual encoder is done on the inverted images, with small learning rates, between $10^{-6}$ and $10^{-4}$, for one single epoch.

\section{Controlled Experiments}

In this section, we perform preliminary controlled experiments to assess the potential of Knowledge Transfer (KT). We aim at learning novel concepts in natural and medical images, and out-of-domain concepts. %

\subsection{Learning novel concepts}
\label{sec:results-new-concepts}

\e{As a proof of concept, we evaluate KT on the RareConcepts dataset, which includes three uncommon classes on natural images (\emph{Moongate, Tonometer, and Gyroscope}) and four classes of fine-grained and deformable categories and \eee{geometric} patterns (i.e. fabric \eee{and parquet} patterns). These were selected by probing CLIP models on web-sourced concepts. We test two CLIP variants (ViT-B/32 and ViT-L/14), %
using the official checkpoints released by OpenAI~\citep{radford2021learning}. The inversion-finetuning setup follows Sec.~\ref{sec:setup}.} %

\eee{Results are shown in Fig.~\ref{fig:kt-results-rare-concepts-all}. We report zero-shot accuracy on target concepts before and after KT. Baseline models show poor recognition of unseen concepts-e.g., CLIP ViT-B/32 fails on Moongate (0\%) and Tonometer. After KT, all models improve on the target classes while largely preserving ImageNet accuracy, indicating minimal forgetting~\citep{kirkpatrick2017overcoming} with a proper choice of learning rate. With proper tuning, CLIP base and large even reach 100\% target accuracy with negligible loss on ImageNet. Overall, these experiments show that KT succesfully introduces unknown concepts and improves existing ones. Detailed results across different learning rate values are reported in the supplementary material, together with an analysis of ViLT.} %

\paragraph{Comparison with other approaches}

\e{To our knowledge, no prior work directly addresses our research question, though some methods use textual information to enhance visual understanding. We qualitatively compare KT with three related directions: (i) improving textual supervision during pre-training (e.g., K-LITE~\citep{shen2022k}), (ii) tuning textual prompts for zero-shot classification (e.g., LLaMP~\citep{zheng2024large}, \cite{menon2023visual}), and (iii) generating synthetic images via text-to-image models (e.g., SynthCLIP~\citep{hammoud2024synthclip}). As shown in Tab.~\ref{tab:comparison-qualitative}, KT uniquely operates without real or synthetic images, requires no generative model, and performs knowledge transfer within the same model.}
\e{Due to these differences, the only reasonable quantitative comparison can be done with the LLM-augmented approaches proposed in~\citep{zheng2024large,menon2023visual}. We compare KT to LLM-augmented prompting methods, using identical captions (see Supplementary Material), in Tab.~\ref{tab:clip-llm}. While both methods succeed on Moongate (likely due to contextual cues such as “garden”), KT performs markedly better on Tonometer and Gyroscope, where LLM prompting even reduces accuracy. This reflects a key difference: KT achieves genuine cross-modal transfer, whereas LLM prompting merely alters text inputs. \ev{For completeness, in our ablation studies (Sec.~\ref{sec:ablation-studies}), we also perform a comparison with a generative setup using Stable Diffusion XL, showing that inverted images outperform synthetic ones, likely because they are explicitly optimized for the target model.}}

\begin{table}
    \centering
    \resizebox{\columnwidth}{!}{
    \begin{tabular}{c >{\centering}p{2cm} *{3}{>{\centering\arraybackslash}p{2cm}}}
    \toprule
         \textbf{Method} & \textbf{Needs real images} & \textbf{Needs Generative Model} & \textbf{Text type} & \textbf{Transfer of knowledge}*\\
    \midrule
         SynthCLIP~\citep{hammoud2024synthclip} & \textcolor{green}{NO} & \textcolor{red}{YES} & Gen. prompt & \textcolor{red}{NO} \\
         K-LITE~\citep{shen2022k} & \textcolor{red}{YES} & \textcolor{green}{NO} & Structured & \textcolor{red}{NO} \\
         LLaMP~\citep{zheng2024large} & \textcolor{red}{YES} & \textcolor{red}{YES} & Gen. prompt & \textcolor{red}{NO}\\
         LLM-aug~\citep{zheng2024large,menon2023visual} & \textcolor{green}{NO} & \textcolor{green}{NO} & Description & \textcolor{red}{NO} \\
         \rowcolor{Apricot} \textbf{KT} (ours) & \textcolor{green}{\textbf{NO}} & \textcolor{green}{\textbf{NO}} & Description & \textcolor{green}{\textbf{YES}} \\
    \bottomrule
    \end{tabular}
    }
    \caption{\textbf{Comparison with existing approaches.}  *refers to transfer within the same model.}
    \label{tab:comparison-qualitative}
\end{table}

\begin{table}
    \centering
    \resizebox{\columnwidth}{!}{%
    \begin{tabular}{l c c c c}
    \toprule
    & \textbf{Moongate} & \textbf{Tonometer} & \textbf{Gyroscope} \\
    \midrule
    CLIP L/14 (base) & 78.95\% & 31.58\% & 90\% \\ 
    CLIP L/14 + LLM~\citep{zheng2024large} & 100\%  & 57.89\% & \textcolor{red}{60\%$\downarrow$}\\
    \rowcolor{Apricot}
    \textbf{CLIP L/14 + KT} (ours) & \textbf{100\%} & \textbf{78.95\%} & \textbf{100\%} \\
    \bottomrule
    \end{tabular}
    }
    \caption{\textbf{Knowledge Transfer (KT) vs LLM-augmented CLIP} (accuracy). Augmenting classification prompts with an LLM can help by introducing contextual cues (e.g. background for moongate), but no transfer of knowledge happens between textual and visual representations. KT provides more reliable improvements.}
    \label{tab:clip-llm}
\end{table}

\subsection{\e{Knowledge Transfer on Medical Imaging}}

Next, we target KT on medical imaging. Medical images are a perfect task for KT, as we can leverage existing medical knowledge in the form of text (e.g. from medical textbooks and encyclopedias) to accurately describe concepts and visual appearance of different pathologies on images such as Chest X-rays (CXR), Computed Tomography (CT) scans, Magnetic Resonance Images (MRI), and Ultrasound images. 
\e{Our experiments use MedCLIP~\citep{wang2022medclip}, a CLIP-based model with BioClinicalBERT~\citep{bioclinicalBERT} as text encoder and Swin Transformer~\citep{liu2021swin} as visual encoder. MedCLIP is pre-trained on MIMIC-CXR~\citep{johnson2019mimic} and CheXpert~\citep{irvin2019chexpert}, containing CXR images and radiological reports covering concepts such as Atelectasis, Cardiomegaly, Consolidation, Edema, Pleural Effusion, and others.
We introduce two new concepts, benign and malignant nodules, into MedCLIP following the same KT protocol as for CLIP, and evaluate zero-shot accuracy on the external JSRT dataset~\citep{shiraishi2000development}.}

Results are reported in Tab.~\ref{tab:kt-results-medclip-lr}, with captions listed in the Supplementary Material. %
As before, we monitor the zero-shot accuracy on previous knowledge using CheXpert-5x200c, to spot instances of catastrophic forgetting. %
KT improves malignant nodule detection from 83.93\% to 92.86\% while retaining comparable results on the source dataset CheXpert-5x200c. For benign nodules, accuracy remains unchanged, likely due to less distinctive visual cues.
Interestingly, we observe a slight overall gain on the source dataset, suggesting more robust feature representations.

\begin{table}
    \centering
    \resizebox{1.0\columnwidth}{!}{%
    \begin{tabular}{l l c c}
        \toprule
         \textbf{Concept} & & \textbf{Baseline} & \textbf{KT} \\
         \midrule
         Benign Nodule & Target Acc. & 54.55\% & \textbf{54.55\%} \\
         & CheXpert 0-shot & 62.10\% & \textbf{62.30\%} \\
         \midrule
         Lung Cancer & Target Acc. & 83.93\% & \textbf{92.86\%} \\
         & CheXpert 0-shot & 62.10\% & \textbf{61.50\%} \\
         \bottomrule
    \end{tabular}
    }
    \caption{\textbf{KT on MedCLIP on the JSRT dataset} (accuracy). The model successfully learns the novel concept of malignant nodules (lung cancer) on CXR images. Benign nodules, on the other hand, are harder to visually differentiate from other findings in CXRs.}
    \label{tab:kt-results-medclip-lr}
\end{table}

\begin{table}
    \centering
    \resizebox{1.0\linewidth}{!}{
    \begin{tabular}{l l l l l l l l}
    \toprule
     & \textbf{Model} & \textbf{Atelect.} & \textbf{Cardiom.} & \textbf{Cons.} & \textbf{Edema} & \textbf{P. Effusion} & \textbf{Top-1} \\
    \midrule
     MedCLIP & \emph{Reference} & 49\% & 69.50\% & 32.50\% & 75.50\% & 84\% & 62.10\% \\
    \midrule
     CLIP (B) & Baseline & 0\% & 2.5\% & 0\% & 0\% & \textbf{94.50\%} & 19.40\% \\
     \rowcolor{Apricot} & \textbf{KT} & 0\% & \textbf{21.5\%} & 0\% & 0\% & 85\% & \textbf{21.30\%} \\
    \midrule
     CLIP (L) & Baseline & \textbf{59.50\%} & 16.50\% & 0\% & 0\% & 35.50\% & 22.40\% \\
    \rowcolor{Apricot} & \textbf{KT} & 4\% & \textbf{32.5\%} & 0\% & 0\% & \textbf{92.5\%} & \textbf{25.90\%} \\
    \bottomrule
    \end{tabular}
    }
    \caption{\textbf{Learning out-of-domain concepts} (natural images $\rightarrow$ medical) shows potential. Accuracy on CheXpert-5x200c.}
    \label{tab:kt-results-clip-chexpert}
\end{table}

\subsection{Out of domain Knowledge Transfer}
Lastly, we assess the potential of KT to introduce novel concepts outside of the training domain. Specifically, we aim to introduce medical concepts into a model trained on natural images. For this purpose, we fine-tune a CLIP model on all five CheXpert classes (atelectasis, cardiomegaly, consolidation, edema, and pleural effusion).
The results are reported in Tab.~\ref{tab:kt-results-clip-chexpert}. We report the performance of MedCLIP as a reference for a model trained on CheXpert. Looking at the top-1 accuracy, we achieve improved results with both versions of CLIP, with the large variant scoring a higher increase from 22.40\% to 25.90\%. However, breaking down the accuracy per class reveals that \emph{i.}) classes with a starting accuracy of 0\% did not improve, and \emph{ii.}) performance in some classes got worse (i.e. pleural effusion and atelectasis). This may be due to the domain gap between the prior knowledge of the model (natural images) and the features specific to the medical domain. Nevertheless, considering this limitation, KT shows potential in zero-shot out-of-domain generalization.

\section{Real-world Experiments}
\label{sec:experiments}

With an extensive evaluation on different datasets and domains, we aim to thoroughly evaluate the potential of KT. 
Here, we focus on improving zero-shot downstream tasks on real-world scenarios, on both novel and known concepts. Namely, we target segmentation, image-text retrieval, and captioning. The detailed description about the experimental setup can be found in the supplementary material.

\subsection{Captioning}

\begin{table}[b]
    \centering
    \resizebox{\columnwidth}{!}{%
    \begin{tabular}{l c c c c}
    \toprule
    & \multicolumn{4}{c}{\textbf{MSCOCO (5K)}} \\
    \textbf{Model} & \textbf{BLEU@4} &\textbf{METEOR} & \textbf{CIDEr} & \textbf{SPICE} \\ 
    \midrule
    CLIP-ViL~\citep{shen2022how} & 40.2 & 29.7 & 134.2 & 23.8 \\
    BLIP~\citep{li2022blip} & 40.4 & - & 136.7 & - \\
    VinVL~\citep{zhang2021vinvl} & 41.0 & 31.1 & 140.9 & \textbf{25.4} \\
    SimVLM~\citep{wang2022simvlm} & 40.6 & 33.7 & 143.3 & 25.4 \\
    LEMON~\citep{hu2021scaling} & \textbf{41.5} & 30.8 & 139.1 & 24.1 \\
    CoCa~\citep{yu2022coca} (proprietary) & 40.9 & \textbf{33.9} & \textbf{143.6} & 24.7 \\
    \midrule
    CoCa & 6.9 & 12.8 & 31.1 & 9.1 \\
    \rowcolor{Apricot} \textbf{CoCa + KT} & \textbf{17.9} & \textbf{19.4} & \textbf{60.8} & \textbf{13.7} \\
    \midrule 
    CoCa FT  & 34.9 & 29.7 & 123.1 & \textbf{23.5} \\
    \rowcolor{Apricot} \textbf{CoCa FT + KT} & \textbf{35.2} & \textbf{29.8} & \textbf{124.0} & 23.3 \\
    \bottomrule
    \end{tabular}}
    \caption{\textbf{Image captioning on MSCOCO}. 
    CoCa refers to the baseline model pre-trained on LAION-2B~\citep{schuhmann2022laionb}, while CoCa FT refers to the model fine-tuned for captioning on MSCOCO. We highlight in bold the best results and the improvements by Knowledge Transfer.
    }
    \label{tab:mscoco-captioning-results}
\end{table}

\begin{table}[t]
    \centering
    \begin{tabular}{p{1.1cm} p{1.93cm} p{1.93cm} p{1.93cm}}
    \toprule
    \footnotesize{\textbf{Sample}} &
    \raisebox{-0.9\totalheight}{\includegraphics[width=60px,height=60px]{img/coco/COCO_val2014_000000581482.jpg}} &
    \raisebox{-0.9\totalheight}{\includegraphics[width=60px,height=60px]{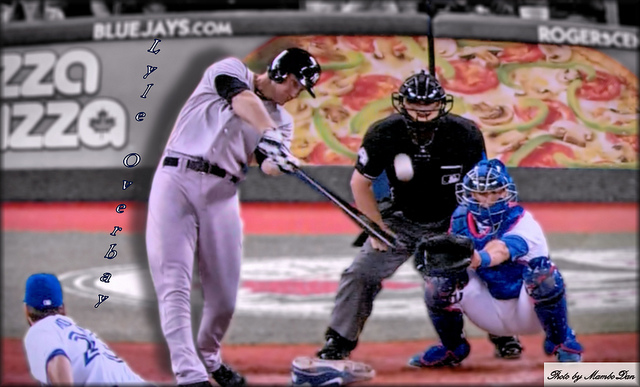}} &
    \raisebox{-0.9\totalheight}{\includegraphics[width=60px,height=60px]{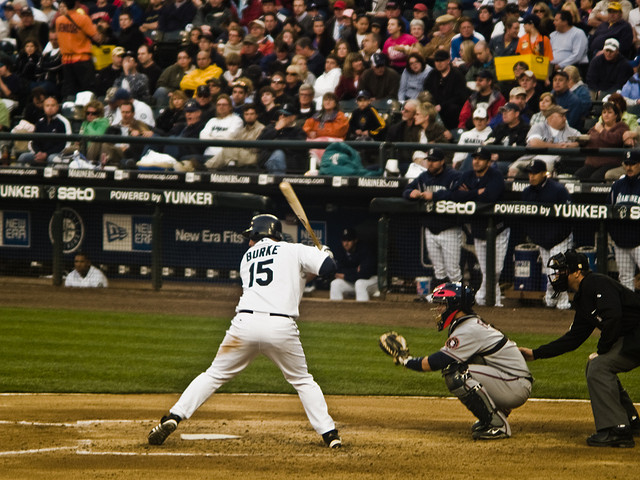}} \\
    \midrule
    {\footnotesize\textbf{Actual Caption}} &
    {\scriptsize A shot of a clock in the train station.} &
    {\scriptsize A baseball player hitting a ball in a professional game.} &
    {\scriptsize A baseball player holding a bat during a baseball game.} \\
    \midrule
    {\footnotesize\textbf{Baseline (CoCa)}} &
    {\scriptsize grand - central - station - new - york.jpg (METEOR 0.0)} &
    {\scriptsize Aaron judge 2016 new york Yankees (METEOR 5.0)} &
    {\scriptsize 20080419 mariners 0001 | by Mike. Smith (METEOR 4.2)} \\
    \midrule
    \rowcolor{Apricot}{\footnotesize\textbf{CoCa+KT}} &
    {\scriptsize A black and white photo of a clock at Grand Central terminal. (METEOR 92.9)} &
    {\scriptsize A baseball player swings his bat at a batter. (METEOR 83.6)} &
    {\scriptsize A baseball player takes a swing at a pitch. (METEOR 86.9)} \\
    \bottomrule
    \end{tabular}
    \caption{\textbf{Visual examples of captioning on MSCOCO.} Illustrative cases where KT improves captioning, with METEOR scores in parentheses.}
    \label{fig:mscoco-captioning-examples-short}
\end{table}

We perform experiments on captioning on the MSCOCO dataset. 
For this task, we employ the CoCa architecture~\citep{yu2022coca}, which is a state-of-the-art captioner. Specifically, we employ the open-source version released by LAION~\citep{ilharco_gabriel_2021_5143773} %
as the original one is proprietary. CoCa is built by adding an autoregressive text decoder to a CLIP model, thus when fine-tuning we apply the InfoNCE loss %
jointly with a captioning loss~\citep{yu2022coca} which aims at predicting the next token $y_t$ given the previous tokens $y_{<t}$ and the image $x$. As captions, we utilize a simple set of templates such as ``\emph{A photo of a X}'', containing different alterations. They are listed in the supplementary material. %

Results are presented in Tab.~\ref{tab:mscoco-captioning-results}. We report different evaluation metrics (BLEU, METEOR, CIDEr, SPICE) computed using the standard pycocoevalcap package~\citep{pycocoevalcap}. %
We perform experiments with two variants of CoCa: one pre-trained on LAION-2B~\citep{schuhmann2022laionb}, and one further fine-tuned (FT) for captioning on MSCOCO. We also report reference results from proprietary CoCa~\citep{yu2022coca} for comparison, along with other methods. With KT, we improve on CoCa FT across almost all metrics, reaching a BLEU@4 of 35.2.
A notable result is achieved with the pre-trained only CoCa, where we improve all metrics by a large margin, sometimes even doubling them (e.g. BLEU@4 from 6.9 to 17.9). We want to point out again that this model is not originally trained for captioning on MSCOCO, and the improvement is introduced by KT alone, without using any real image at all. %
We report some visual results in Tab.~\ref{fig:mscoco-captioning-examples-short}, where the improvement is notable; more examples can be found in the supplementary material.

\begin{table*}
    \centering
    \resizebox{\linewidth}{!}{
    \begin{tabular}{l c c c c c c c c c c c c}
    \toprule
    &  \multicolumn{3}{c}{\textbf{Lung Nodules$^\dagger$}} & \multicolumn{3}{c}{\textbf{Lung Pneumothorax}$^\dagger$} & \multicolumn{3}{c}{\textbf{Breast Ultrasound}} & \multicolumn{3}{c}{\textbf{Brain MRI}}  \\
    \cmidrule(l){2-4} \cmidrule(l){5-7} \cmidrule(l){8-10} \cmidrule(l){11-13} 
    \textbf{Model} & \textbf{DSC} & \textbf{NSD} & \textbf{IoU} & \textbf{DSC} & \textbf{NSD} & \textbf{IoU} & \textbf{DSC} & \textbf{NSD} & \textbf{IoU} & \textbf{DSC} & \textbf{NSD} & \textbf{IoU} \\
    \midrule
    MedCLIP-SAMv2 & \textbf{14.83}\% & 17.30\% & 8.64\% & 6.30\% & 7.61\% & 3.75\% 
    & 56.25\% & 59.44\% & 47.81\% & 17.20\% & 20.97\% & 12.05\% \\
    \rowcolor{Apricot} \textbf{KT} (1e-5) & 13.95\% & 17.45\% & 8.75\% & 6.28\% & 7.59\% & 3.77\% & \textbf{58.23}\% & \textbf{61.56}\% & \textbf{49.52}\%  & 15.90\% & 19.36\% & 11.10\% \\
    \rowcolor{Apricot} \textbf{KT} (2e-5) & 14.10\% & 17.65\% & 8.83\% & \textbf{6.41}\% & \textbf{7.76}\% & \textbf{3.83}\% & 54.36\% & 57.30\% & 46.30\% & \textbf{18.13}\% & \textbf{22.26}\% & \textbf{12.62}\% \\
    \rowcolor{Apricot} \textbf{KT} (1e-4) & 14.35\% & \textbf{18.03}\% & \textbf{9.04}\% & 6.02\% & 7.29\% & 3.59\% & - & - & - & - & - & -\\
    \bottomrule
    \end{tabular}
    }
    \caption{\textbf{Improvements in zero-shot segmentation}. $^\dagger$ denotes novel concepts that are not included in the original MedCLIP-SAMv2 training data~\citep{koleilat2024medclip}. Prompts used for segmentation are reported here: \textbf{P1} A medical chest CT scan showing circular spots of varying size within the lungs, suggesting either benign or malignant nodules; \textbf{P2} A medical chest x-ray showing an abnormal collection of air within the pleural cavity, suggesting a pneumothorax; \textbf{P3} A medical breast mammogram showing an irregularly shaped, spiculated mass suggestive of a malignant breast tumor; \textbf{P4} A brain MRI showing a bright or dark mass with irregular edges suggestive of a brain tumor or glioma.}
    \label{tab:medclip-samv2-results}
\end{table*}

\begin{figure}[!h]
    \centering
    \resizebox{1.0\columnwidth}{!}{%
        \begin{subfigure}[t]{\linewidth}
            \centering
            \includegraphics[width=0.22\linewidth]{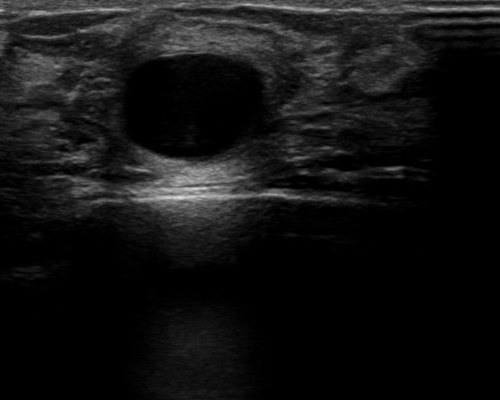}
            \includegraphics[width=0.22\linewidth]{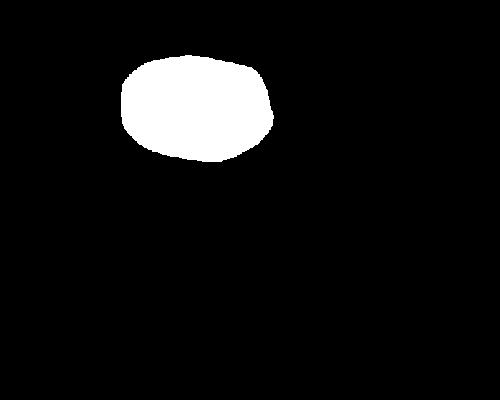}
            \includegraphics[width=0.22\linewidth]{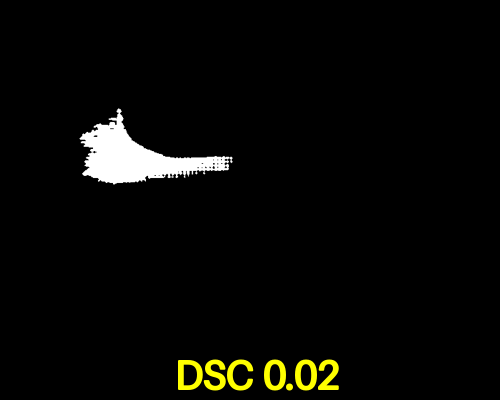}
            \includegraphics[width=0.22\linewidth]{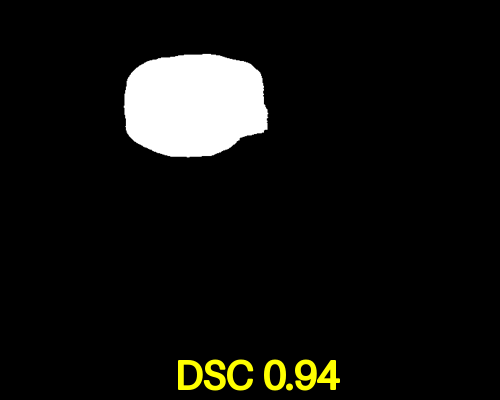} \\[2pt]
            \includegraphics[width=0.22\linewidth]{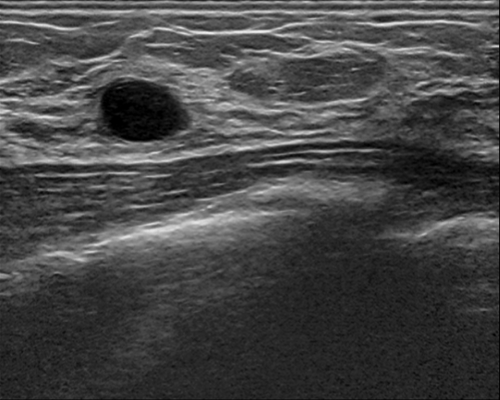}
            \includegraphics[width=0.22\linewidth]{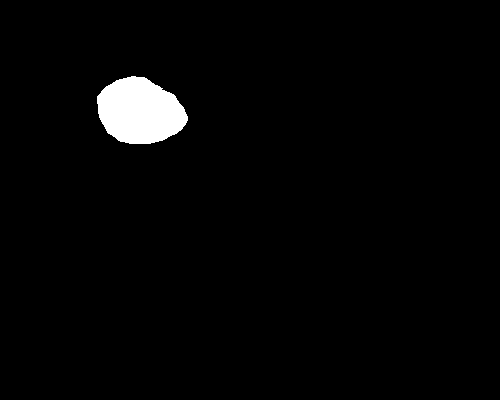}
            \includegraphics[width=0.22\linewidth]{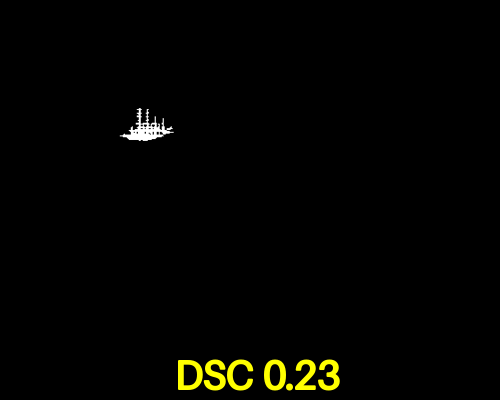}
            \includegraphics[width=0.22\linewidth]{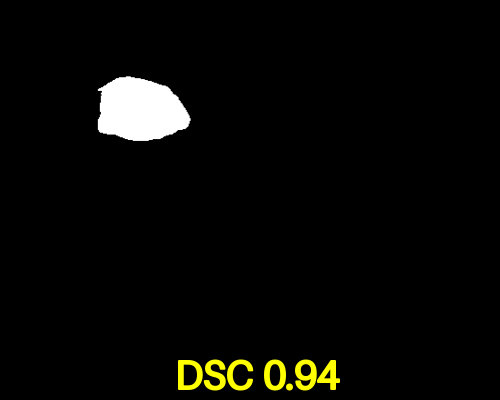} \\[2pt]
            \begin{subfigure}[t]{0.22\linewidth}
                \includegraphics[width=\linewidth]{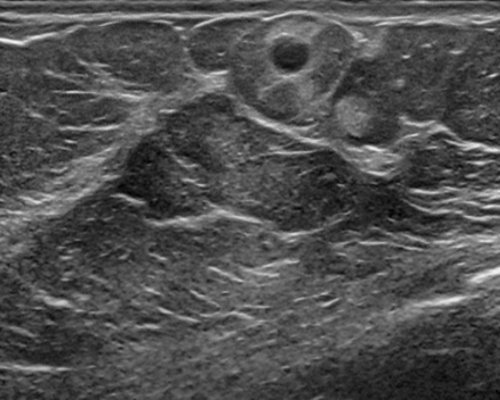}
                \caption*{Image}
            \end{subfigure}
            \begin{subfigure}[t]{0.22\linewidth}
                \includegraphics[width=\linewidth]{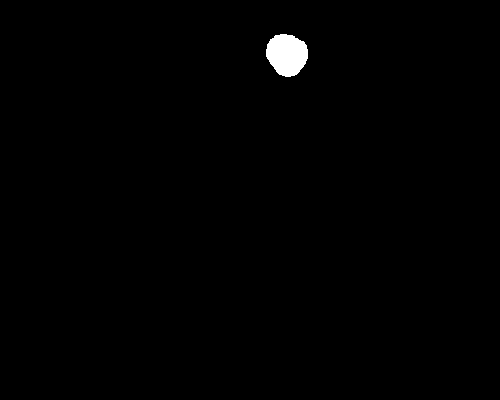}
                \caption*{Ground Truth}
            \end{subfigure}
            \begin{subfigure}[t]{0.22\linewidth}
                \includegraphics[width=\linewidth]{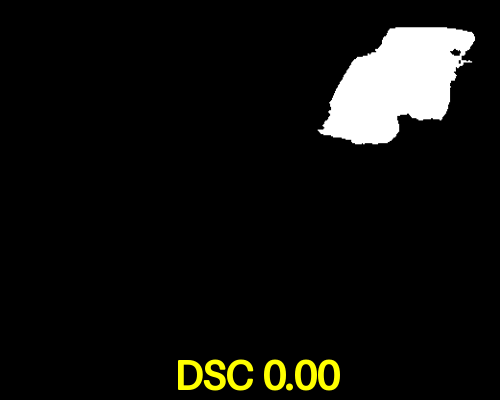}
                \caption*{MedCLIP-SAMv2}
            \end{subfigure}
            \begin{subfigure}[t]{0.22\linewidth}
                \includegraphics[width=\linewidth]{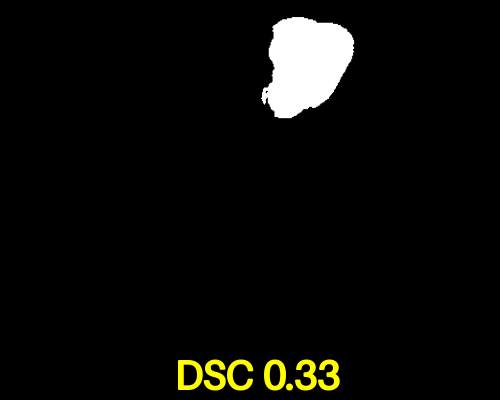}
                \caption*{\textbf{KT}}
            \end{subfigure}
        \end{subfigure}
    } %
    \caption{\textbf{Qualitative evaluation of KT on breast tumor segmentation} (UDIAT dataset). We report illustrative examples where knowledge transfer improved segmentation, in terms of DSC.}
    \label{fig:segmentation-breast}
\end{figure}

\begin{table}
    \centering
    \resizebox{\columnwidth}{!}{%
    \begin{tabular}{@{}lcccccc@{}}
    \toprule
    & \multicolumn{6}{c}{\textbf{Flickr30k (1K)}} \\
    & \multicolumn{3}{c}{\textbf{Text Retrieval}} & \multicolumn{3}{c}{\textbf{Image Retrieval}} \\ 
    \cmidrule(l){2-4} \cmidrule(l){5-7} 
    \textbf{Model} & \textbf{R@1} & \textbf{R@5} & \textbf{R@10} & \textbf{R@1} & \textbf{R@5} & \textbf{R@10} \\ 
    \midrule
    ViLBERT \citep{lu2019ViLBERT} & - & -  & - & 31.9\% & 61.1\% & 72.8\% \\
    Unicoder-VL \citep{li2020unicoder} & 64.3\% & 85.8\% & 92.3\% & 48.4\% & 76.0\% & 85.2\% \\
    ImageBERT \citep{qi2020imagebert} & 70.7\% & 90.2\% & 94.0\% & 54.3\% & 79.6\% & 87.5\% \\
    ViLT-B/32 (original) \citep{kim2021vilt} & 73.2\% & 93.6\% & 96.5\% & 55.0\% & 82.5\% & 89.8\% \\
    \midrule
    ViLT-B/32 (huggingface) & 73.8\% & 93.5\% & 96.5\% & 57.3\% & 83.9\% & 90.4\% \\
    \rowcolor{Apricot} \textbf{ViLT-B/32 + KT} (lr 9e-7) & \textbf{74.6\%} & \textbf{93.8\%} & 96.4\% & \textbf{57.8\%} & \textbf{84.0\%} & \textbf{90.5\%} \\
    \rowcolor{Apricot} \textbf{ViLT-B/32 + KT} (lr 2e-6) & \textbf{74.6}\% & 93.7\% & \textbf{96.5\%} & \textbf{57.8}\% & \textbf{84.0}\% & \textbf{90.5}\% \\ 
    \bottomrule
    \end{tabular}%
    }
    \caption{\textbf{Text and image retrieval} on Flickr30k. Recall scores are shown at top 1, 5 and 10 levels. Our results are based on huggingface's ViLT. Original results and other comparisons from~\citep{kim2021vilt}.}%
    \label{tab:flickr30k-eval}
\end{table}

\subsection{Segmentation}
For segmentation, we employ the zero-shot method MedCLIP-SAMv2~\citep{koleilat2024medclip, koleilat2024medclipsamv2}. It works by computing activation maps from a pre-trained CLIP model, and using them as query for the Segment Anything Model (SAM)~\citep{kirillov2023segment}.
Activation maps are computed using Multi-Modal Information Bottleneck Attribution (M2IB)~\citep{wang2023visual}, using a target image and a query prompt. 
Here, we aim at improving the quality of the activation maps on different concepts by leveraging KT. This, in turn, should result in a higher accuracy of the final segmentation.
We target four different segmentation tasks: lung nodules segmentation on CT images (UnitoChest), pneumothorax segmentation on CXR images (SIIM Pneumothorax), breast nodule segmentation on ultrasound images (UDIAT), and glioma segmentation in MRIs (BraTS23). 

The overall results across all segmentation tasks are presented in Tab.~\ref{tab:medclip-samv2-results}. The captions used for inversion are reported in the supplementary material. To compute the M2IB activation maps on the fine-tuned models, we employ descriptive prompts as suggested in~\citep{koleilat2024medclipsamv2}. The prompts are reported in Tab.~\ref{tab:medclip-samv2-results} as P1 to P4 for each task. We also report reference results of MedCLIP-SAMv2 on each task. Compared to the original setting of MedCLIP-SAMv2, lung nodules and lung pneumothorax are completely novel concepts. There is also a slight difference in the brain glioma class compared to the original brain tumor task, explained in the supplementary file. We employ three metrics to assess the segmentation quality, namely the Dice-Sørensen Coefficient (DSC), Normalized Surface Distance (NSD), and Intersection over Union (IoU). We report results with different values of fine-tuning learning rate. We can observe an increase in segmentation metrics across all tasks, notably in breast ultrasound (NSD 59.44\% to 61.56\%) and brain MRIs (NSD 20.97\% to 22.26\%). For lung nodules and pneumothorax, the improvement is less pronounced, probably because the novelty of the task makes improving more difficult in the MedCLIP-SAM setting. We report some visual examples on breast tumor segmentation in Fig.~\ref{fig:segmentation-breast}, showcasing the improvements of KT.

\subsection{Text and image retrieval}
We perform experiments on text and image retrieval on the Flickr30k dataset. For these experiments, we employ the huggingface version %
of ViLT~\citep{kim2021vilt}. 
To fine-tune ViLT with KT, we employ captions of common concepts that may help improve the model's general knowledge. For this purpose, we use the 80 object categories from MSCOCO as target concepts, using the method presented in Sec.~\ref{sec:setup}, using ChatGPT-4. All captions are reported in the supplementary material. %
For each caption, we generate 10 inverted images, for a total of 800 inverted images. Fine-tuning is performed as in Sec.~\ref{sec:results-new-concepts}, by maximizing the ITM score for positive pairs and minimizing it for negative ones. %

The results of zero-shot text and image retrieval are reported in Tab.~\ref{tab:flickr30k-eval}. For comparison, we also report the original results of ViLT from~\citep{kim2021vilt}, alongside other relevant baselines. Results are shown in terms of recall (marked as R) computed at different levels (top-1, top-5, and top-10 recall). As we observe from the results, KT consistently improves the results across all metrics for both image and text retrieval tasks. Notably, we score an improvement of almost 1\% from 73.8\% to 74.6\% on the text retrieval task. Additional configurations are available in the supplementary material.

\subsection{Ablation studies}
\label{sec:ablation-studies}

\begin{figure}
    \centering
    \includegraphics[width=\linewidth]{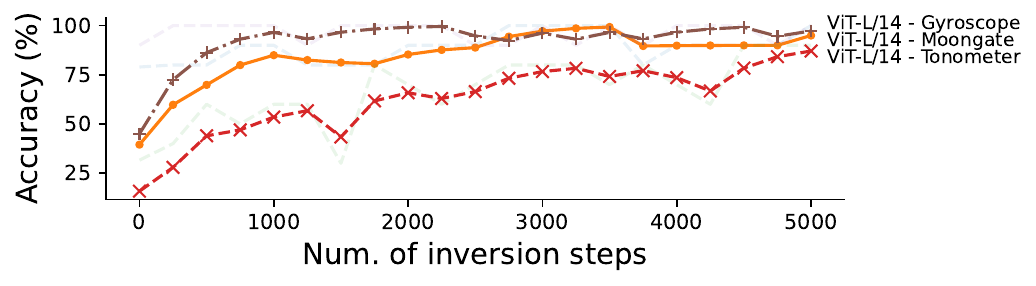}
    \caption{\textbf{KT results at different inversion steps.} Better inversion quality (more steps) generally leads to improved results.}
    \label{fig:inversion-ablation-steps}
\end{figure}

\begin{figure}
    \centering
    \begin{subfigure}[t]{0.49\linewidth}
        \includegraphics[width=\linewidth]{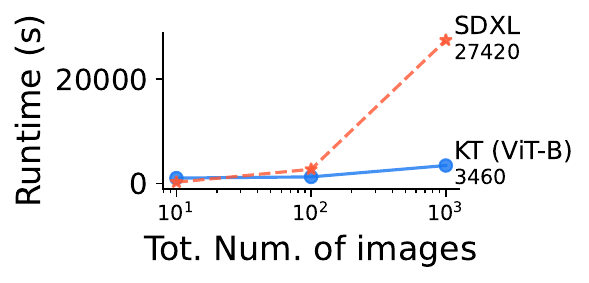}
    \end{subfigure}
    \begin{subfigure}[t]{0.49\linewidth}
        \includegraphics[width=\linewidth]{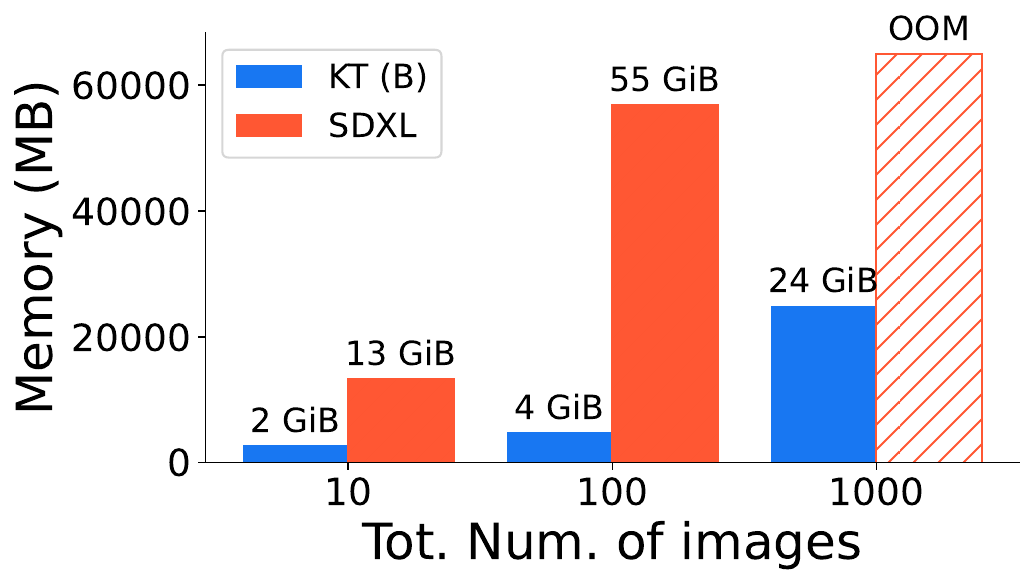}
    \end{subfigure}
    \caption{\textbf{Compute and scaling of inversion} \emph{(left)} total runtime \emph{(right)} memory required to generate 10, 100, and 1k images.}
    \label{fig:inversion-scaling}
\end{figure}

\e{In this section, we analyze the main factors influencing KT's performance using the RareConcepts dataset, namely inversion hyperparameters and quality of the textual description. We also report computational details and scaling of inversion, compared to other approaches such as generative models. Additional ablations on fine-tuning strategy and caption construction are provided in the Supplementary Material.} \\
\noindent\e{\textbf{Inversion hyperparameters} Figure~\ref{fig:inversion-ablation-steps} and Table~\ref{tab:inversion-hyperparameters-ablation} summarize how inversion quality affects downstream accuracy. Increasing the number of inversion steps generally improves reconstruction fidelity, resulting in higher KT accuracy (Fig.~\ref{fig:inversion-ablation-steps}). Similarly, applying image augmentation and total variation regularization during inversion leads to more stable and accurate representations (Tab.~\ref{tab:inversion-hyperparameters-ablation}). The full configuration combining both techniques achieves the best accuracy.}\\
\noindent\e{\textbf{Description quality} Tab.~\ref{tab:ablation-caption-quality-sdxl} evaluates KT’s robustness to caption length and quality. KT consistently performs well across short (P1), medium (P2), and long (P3) captions, showing only minor variation. Human-written captions with concise, relevant descriptions slightly outperform LLM-generated or verbose ones, which may also contain irrelevant text.  Overall, KT remains stable across prompt variations. The captions used are listed in the supplementary material. } \\
\noindent\e{\textbf{Model inversion vs Stable Diffusion}}
\ev{In this ablation, we employ a generative approach based on SDXL using the same prompts of previous experiments (P3) on the RareConcepts dataset, to obtain images used for fine-tuning. As summarized in the last column of Tab.~\ref{tab:ablation-caption-quality-sdxl}, KT outperforms the generative SDXL baseline (average 88.3\% vs. 86.7\%), while requiring much less compute and memory.} \\
\noindent\e{\textbf{Compute efficiency and scaling}
As shown in Fig.~\ref{fig:inversion-scaling}, our inversion-based KT scales efficiently with dataset size. Compared to SDXL, KT requires substantially less compute and memory (e.g., 3.5k s vs. 27k s for 1k images), remaining practical even for large-scale applications.}

\begin{table}
    \centering
    \resizebox{\linewidth}{!}{%
    \begin{tabular}{c c c c c c c}
    \toprule
    \textbf{Name} & \textbf{Augmentation} & \textbf{TV Strength} ($\alpha$) & \textbf{Moongate} & \textbf{Tonometer} & \textbf{Gyroscope} & \textbf{Avg.} \\
    \midrule
    \emph{Baseline} & - & - & 0\% & 50\% & 90\% & 46.66\%\\
    A1  & No & 0 & 20\% & 80\% & 100\% & 66.67\% \\
    A2 & Yes & 0.001 & 30\% & 80\% & 100\% & 70\% \\
    \midrule
    {Full} & Yes & {0.005} & \textbf{60\%} & \textbf{80\%} & \textbf{100\%} & \textbf{80\%}\\
    \bottomrule
    \end{tabular}
    }
    \vspace{2pt}
    \caption{\textbf{Inversion hyperparameters on CLIP (B).} Augmentation and regularization improve quality, achieving better overall results. %
    }
    \label{tab:inversion-hyperparameters-ablation}
\end{table}

\begin{table}
    \centering
    \resizebox{\linewidth}{!}{%
    \begin{tabular}{c c c c c c |c c}
    \toprule
    \textbf{Model} & \textbf{Concept} & \textbf{Baseline} & \textbf{P1} & \textbf{P2} & \textbf{P3} & \textbf{SDXL}\\
    \midrule
     \multirow{3}{*}{CLIP ViT-B/32} & Moongate & 0\% & 50\% & 70\% & 60\% & 70\% \\
      & Tonometer & 50\% & 80\% & 80\% & 80\% & 70\% \\
      & Gyroscope & 90\% & 100\% & 100\% & 100\% & 90\% \\
    \midrule
    \multirow{3}{*}{CLIP ViT-L/14} & Moongate & 78.95\% & 100\% & 100\% & 100\% & 90\%\\
      & Tonometer & 31.58\% & 100\% & 90\% & 90\% & 100\%\\
      & Gyroscope & 90\% & 100\% & 100\% & 100\% & 100\% \\
    \midrule
     Avg &  & 56.76\% & 88.33\% & 90\% & 88.33\% & 86.67\%\\
    \bottomrule 
    \end{tabular}
    }
    \vspace{2pt}
    \caption{\textbf{Robustness of KT to prompt variations and comparison with SDXL} (\textbf{P1:} short human caption < 8 words; \textbf{P2:} mid-size human caption < 16 words; \textbf{P3:} longer LLM caption > 32 words).}
    \label{tab:ablation-caption-quality-sdxl}
    \vspace{-1.5em}
\end{table}

\section{Conclusions and Future Works}

We present a way to learn novel visual concepts by only using their textual descriptions, with a method we call Knowledge Transfer (KT). Through extensive evaluation, we show that KT can successfully introduce novel concepts in pre-trained VLMs, without hurting performance on previous tasks. We also show that KT can improve the results of downstream zero-shot tasks, such as segmentation, text-image retrieval, and captioning, and also shows potential for out-of-domain generalization, for example on medical images. %
The proposed method is based on model inversion to synthesize \emph{ideal} images for a target concept, that are later used to fine-tune the VLM in an image-text matching fashion, such as~CLIP~\citep{radford2021learning}. %
Our method leverages prior knowledge in pre-trained VLMs, with the aim of aligning known visual concepts to novel high-level ones. 

One key design choice in our KT framework is not requiring parameters to be shared between visual and textual encoders, thus making it generally applicable regardless of the specific architecture of the VLM. \eee{We believe that by relaxing this constraint, it should possible to achieve KT with an alternative framework, for example by leveraging multimodal neurons}~\citep{schwettmann2023multimodal} and employing only textual captions (e.g., fine-tuning the VLM with masked language modeling). We provide a brief presentation of such alternative framework in the supplementary material, leaving an in-depth investigation to future work. 

To the best of our knowledge, this is the first work attempting to leverage prior knowledge inside a neural network to teach it novel concepts. We believe this work can pave the way for future research on this topic, especially in data-limited domains where imaging data is scarce but we have access to textual human knowledge, such as medical imaging.

{
    \small
    \bibliographystyle{ieeenat_fullname}
    \bibliography{main}
}

\clearpage
\newpage
\appendix
\appendix

\section{Impact and Limitations}

We believe that Knowledge Transfer has the potential to be an impactful technique for introducing novel concepts in pre-trained models. Overall, Knowledge Transfer is quite cheap in terms of computational requirements, as it works by only fine-tuning on just a handful of synthesized samples. Thus, it is very quick and does not need a large amount of memory. In this sense, it may be comparable to parameter-efficient fine-tuning (PEFT) techniques, such as low-rank adaptation (LoRA)~\citep{hu2021lora}, which minimize the amount of memory required for fine-tuning. However, compared to PEFT, Knowledge Transfer does not require any real data besides a single textual description for each novel concept.

From this point onward, we will refer to the KT algorithm described in the main paper as \textbf{Explicit} Knowledge Transfer, namely the variant that relies on an inversion step to synthesize a visual example before fine-tuning. In contrast, we use the term \textbf{Implicit} Knowledge Transfer to denote approaches that avoid this inversion step and instead rely on shared parameters between modalities (e.g., multimodal neurons), enabling transfer through purely textual objectives such as MLM.

The main limitation of Explicit Knowledge Transfer lies in the inversion step, which takes the most time compared to fine-tuning. If this step could be avoided, we could achieve near real-time learning of novel concepts with minimal computational requirements. This could enable the development of rapidly improving intelligent agents in many real-world applications. We hypothesize this is possible with Implicit Knowledge Transfer, for example by using Masked-Language Modeling (MLM) as a proxy for knowledge transfer. However, in this work, we do not focus on this topic, as preliminary experiments (shown in Sec.~\ref{sec:implicit-transfer-preliminary}) did not achieve satisfactory results compared to Explicit Knowledge Transfer. 

Another limitation lies in the limited comparison with state-of-the-art approaches; however, to the best of our knowledge, we are not aware of other works sharing the same goal as ours.

\section{Knowledge Transfer}

\begin{figure*}[t]
    \centering
    \includegraphics[width=0.9\linewidth]{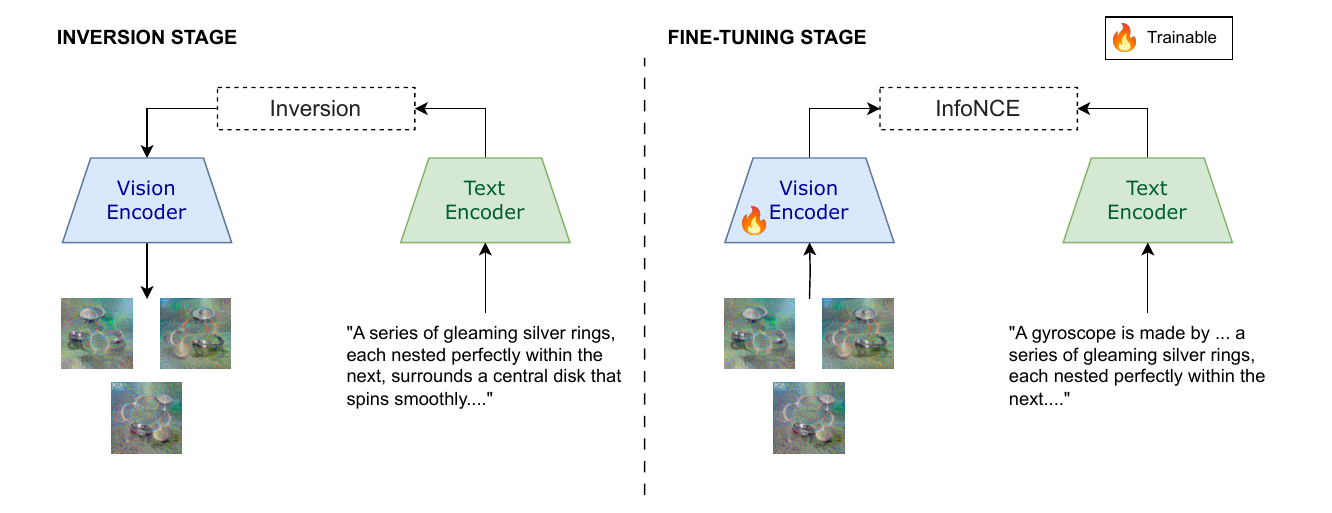}
    \caption{\textbf{Graphical overview of Knowledge Transfer.} Starting from a textual description of the target concept, we synthesize images via model inversion (left) then, using an image-text matching loss, we fine-tune the visual encoder to match the concept (right). In this way, we leverage prior knowledge contained in the model (from pre-training) to learn novel concepts.}
    \label{fig:method-overview-supp}
\end{figure*}

A general overview of explicit Knowledge Transfer can be found in Fig.~\ref{fig:method-overview-supp}.

\subsection{Examples of inverted images}

\begin{figure*}
    \centering
    \begin{subfigure}[t]{0.48\linewidth}
    \includegraphics[width=46px,height=46px]{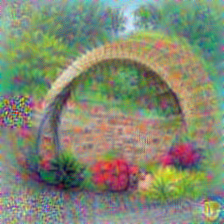}
    \includegraphics[width=46px,height=46px]{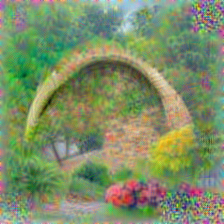}
    \includegraphics[width=46px,height=46px]{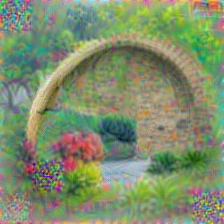}
    \includegraphics[width=46px,height=46px]{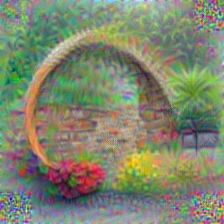}
    \includegraphics[width=46px,height=46px]{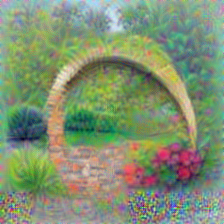}
    \includegraphics[width=46px,height=35px]{img/moongate/1.png}
    \includegraphics[width=46px,height=35px]{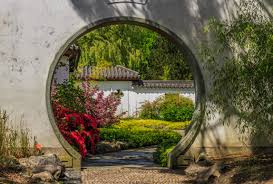}
    \includegraphics[width=46px,height=35px]{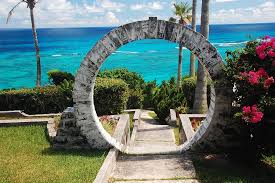}
    \includegraphics[width=46px,height=35px]{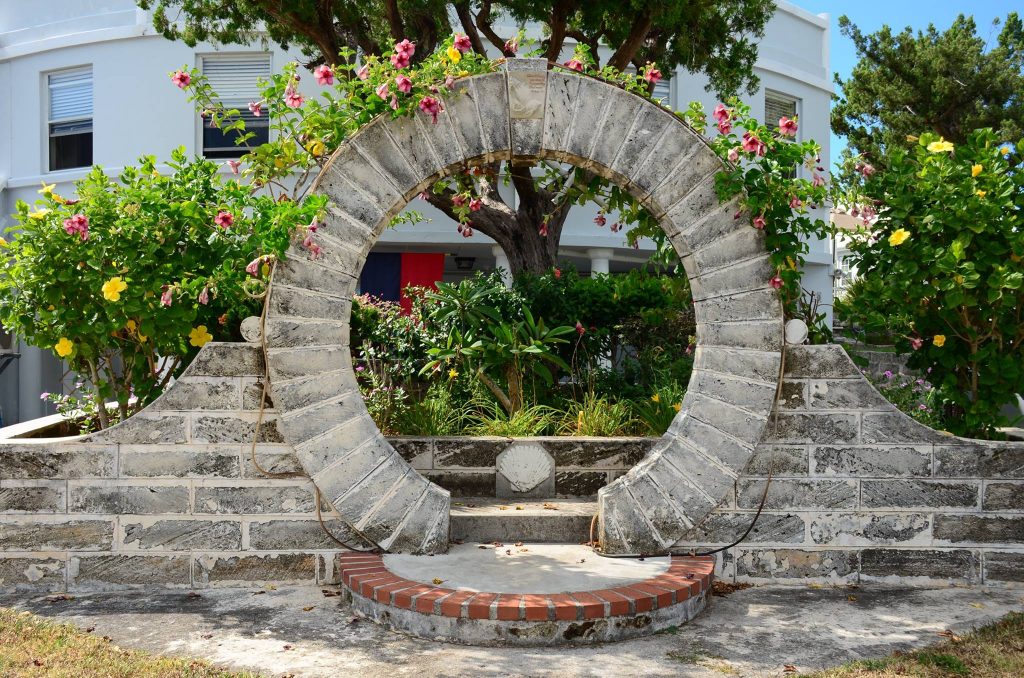}
    \includegraphics[width=46px,height=35px]{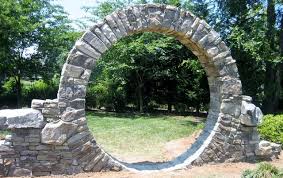}
    \caption{Moongate. Caption: \emph{A perfectly circular archway built from uniformly cut stones or bricks, set into a larger wall. It forms a smooth circle, framing views of gardens or landscapes beyond, creating a picturesque portal.}}
    \end{subfigure}
    \hfill
    \begin{subfigure}[t]{0.48\linewidth}
    \includegraphics[width=46px,height=46px]{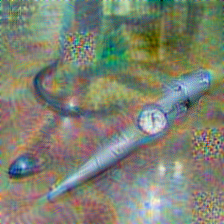}
    \includegraphics[width=46px,height=46px]{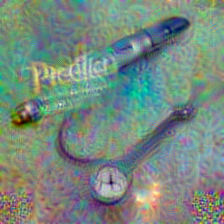}
    \includegraphics[width=46px,height=46px]{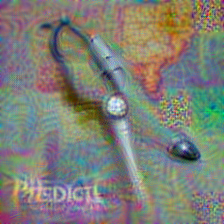}
    \includegraphics[width=46px,height=46px]{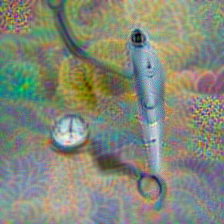}
    \includegraphics[width=46px,height=46px]{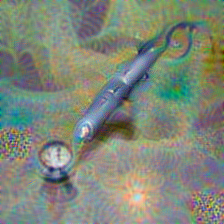}
    \includegraphics[width=46px,height=40px]{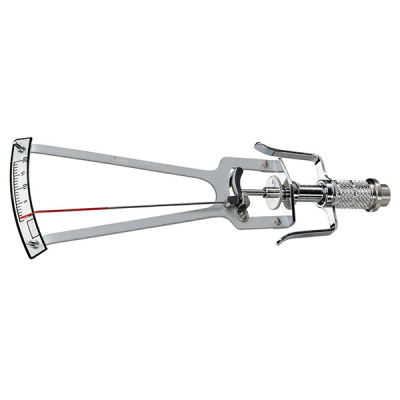}
    \includegraphics[width=46px,height=40px]{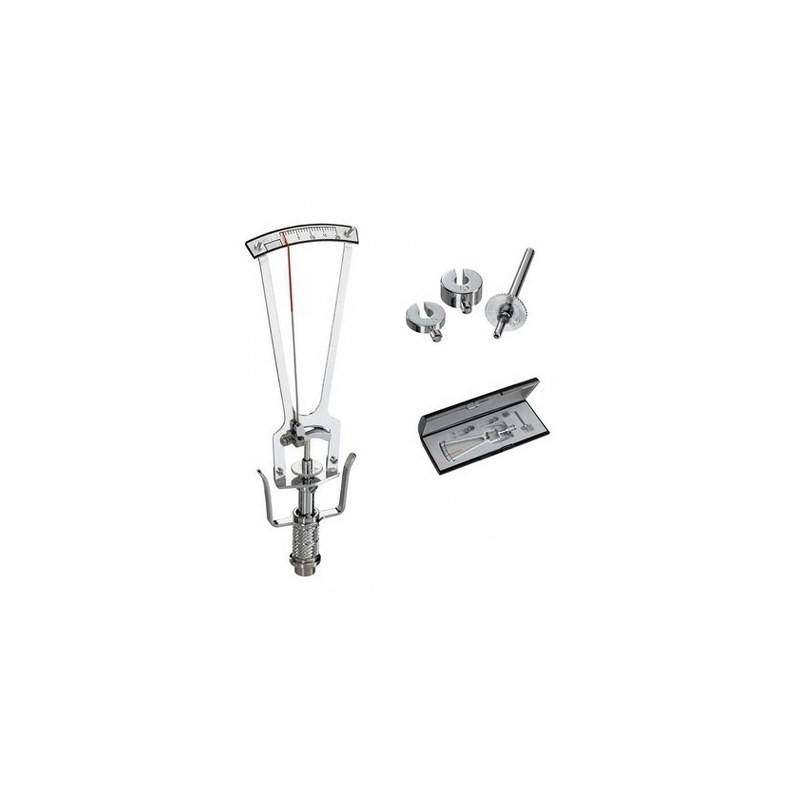}
    \includegraphics[width=46px,height=40px]{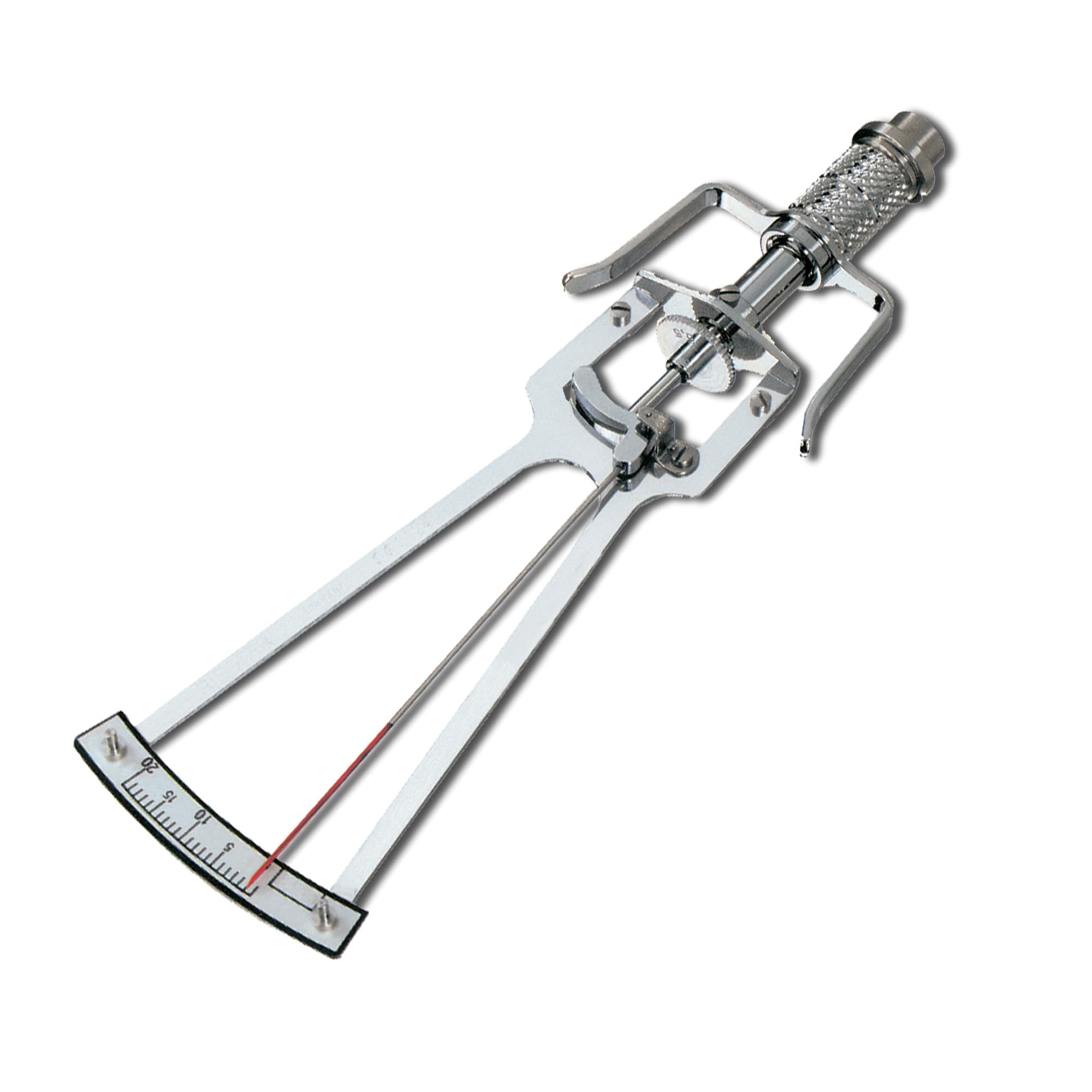}
    \includegraphics[width=46px,height=40px]{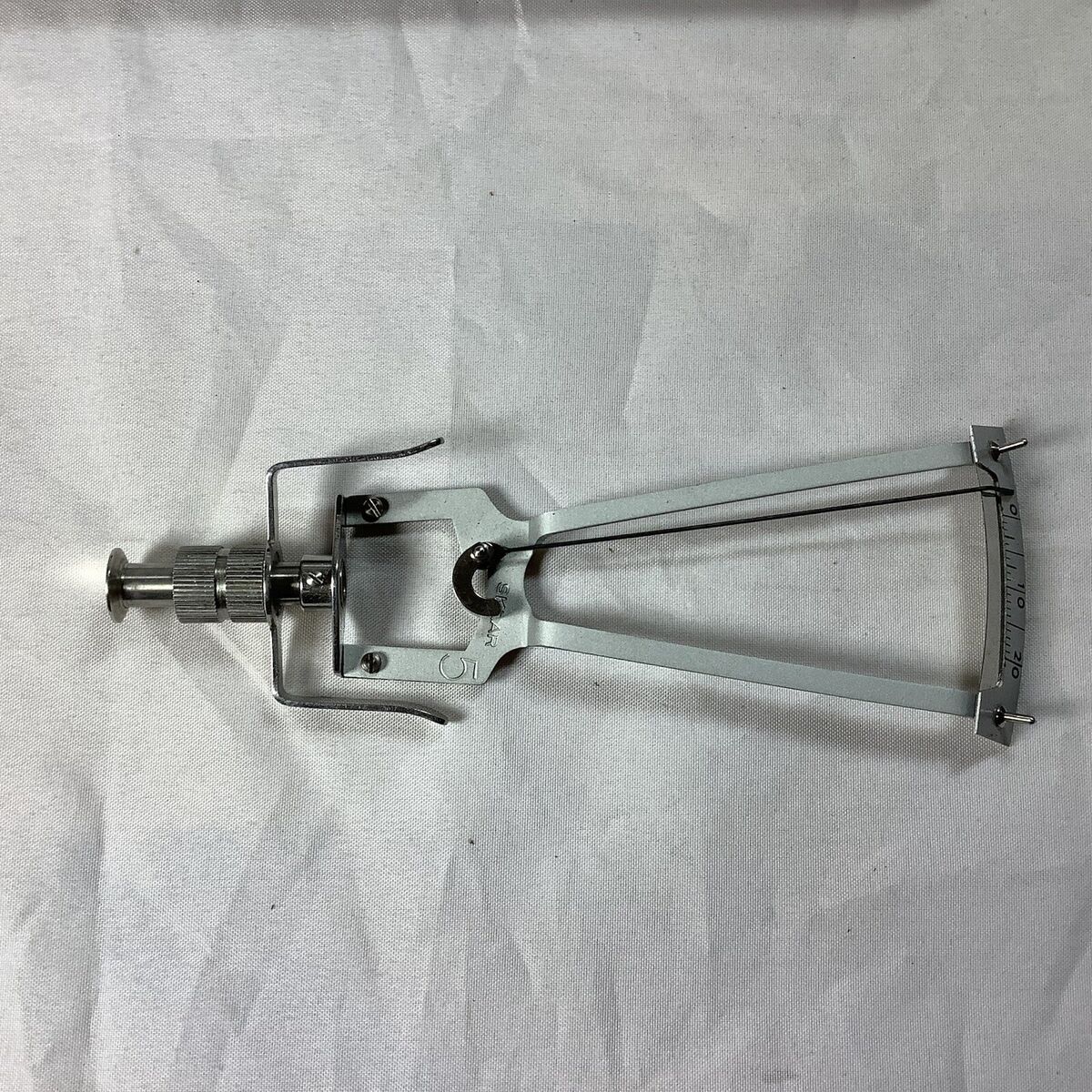}
    \includegraphics[width=46px,height=40px]{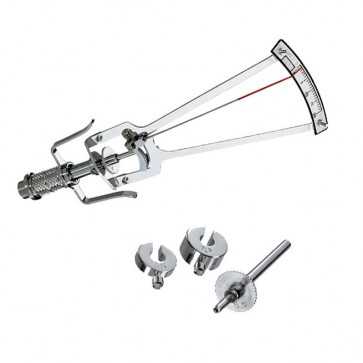}
    \caption{Tonometer. Caption: \emph{A slender, pen-like probe attached to a small base equipped with precise dials and gauges. This tool is often part of a larger medical apparatus, featuring a metallic finish and a refined, professional appearance.}}
    \end{subfigure}
    \caption{Example of inverted images (top) and real images (bottom) from rare concepts that CLIP struggles to classify correctly.}
    \label{fig:inversion-example-full}
\end{figure*}

Examples of inverted images can be found in Fig.~\ref{fig:inversion-example-full}.

\subsection{Possible improvements of Explicit Transfer}

We start from the inversion equation: 

\begin{equation}
\begin{split}
    \hat{X}^*_V = f_V^{-1}(f_T(X_T)) \qquad\qquad\qquad\qquad\qquad\qquad \\
    \approx \max_{\hat{X}^*_V}sim(A(f_V(\hat{X}^*_V)), f_T(X_T)) + \alpha R(\hat{X}^*_V) \quad.
    \label{eq:inversion}
\end{split}
\end{equation}

\subsubsection{Relaxation of Eq.~\ref{eq:inversion}}

Computing $\hat{X}^*_V$ as in Eq.~\ref{eq:inversion} might produce images that are widely different from the training distribution of natural images, as shown in Fig.~\ref{fig:inversion-example-full}. So, instead of inverting the whole visual encoder $f_V$, we can invert just a subset of layers $\Psi_V \subset f_V$, starting from the top of the model:

\begin{equation}
\begin{split}
    \hat{Z}^*_V = \Psi_V^{-1}(f_T(X_T)) \approx \max_{\hat{Z}^*_V}sim(\Psi_V(\hat{Z}^*_V), f_T(X_T)) \\
    + R(\hat{Z}^*_V)
\end{split}
\end{equation}

\noindent where $R$ could be a regularization similar to style transfer~\citep{gatys2016image} to encourage $\hat{Z}^*_V$ to be similar to the intermediate representations of natural images.

\subsection{Implicit Knowledge Transfer}
\label{sec:implicit-knowledge-transfer}
Although in this work we focus on explicit knowledge transfer, we briefly present the idea behind Implicit Knowledge Transfer for the sake of completeness. It has been shown how multi-modal neurons can be found in multi-modal models~\citep{goh2021multimodal, schwettmann2023multimodal}. These neurons exhibit high activation on the same concepts in either modality, meaning that they are able to capture cross-modal representations. We hypothesize that in a shared-parameter architecture (e.g. early-fusion transformers~\citep{mo2024unveiling, kim2021vilt}) it should be possible to exploit these neurons for knowledge transfer, for example with simple masked language modeling on the novel concept description, effectively eliminating the need for model inversion. For this purpose, early-fusion architectures that can process single modalities independently would be required. However, to the best of our knowledge, we are not aware of many large pre-trained models satisfying these requirements at the time being, hence we leave an in-depth exploration of this path for future research. Hints that training on different modalities independently can help can be found in the literature, for example during pre-training of U-VisualBERT~\citep{li2021unsupervised}. Even more relevant to our research, authors in~\citep{wang2022simvlm} report some capabilities of cross-modal transfer on SimVLM, however, the model is proprietary and we are unable to reproduce their claims. Thus, here we focus on ViLT and we report some preliminary analysis in Sec.~\ref{sec:implicit-transfer-preliminary}. 

\subsection{Open questions}

\paragraph{Q1 Domain Gap.} Inverted images, as shown in Fig.~\ref{fig:inversion-example-full}, appear widely different from natural images. However, as shown by the results in the paper, fine-tuning the models on them leads to improved results. Is a domain gap present between inverted and real images? Or is it indicative of a fundamental difference in which deep models process visual information? This phenomenon may be linked with adversarial attacks~\citep{xu2020adversarial}.

\paragraph{Q2 Generalizability of inversion} An interesting point to analyze, which could provide some insights for Q1, is the generalizability of the inverted images. For example, can images inverted with a certain model (e.g. CLIP) be used for training some other model from scratch? Or are they ``fitted'' to only work with the specific model used for inversion?

\paragraph{Q2 Catastrophic Forgetting} To what extent can we prevent catastrophic forgetting when applying Knowledge Transfer? In this work, we show that lower learning rates generally achieve a good trade-off between learning novel concepts and preserving previous information. However, there is still room for improvement. For example, LoRA~\citep{hu2021lora} has been shown to help in avoiding catastrophic forgetting during fine-tuning, hence applying it during Knowledge Transfer could further improve the results. Also, Implicit Transfer (on shared-parameter models) might avoid catastrophic forgetting better than Explicit Transfer, for example by focusing on multi-modal neurons.

\section{Experimental Setup}

\subsection{Training Details}

\paragraph{Captioning}
To produce descriptive captions for the new concepts, we employ a LLM-based approach. Specifically, for natural images, we use Llama-3 Instruct (with 8B parameters)~\citep{llama3modelcard} with the following prompt: \emph{``Generate a small description of the ImageNet class <class name> without using the word itself. The description must contain visual cues useful for recognizing the subject with low-level and accurate details. Please don't insert anything else in the response except the description.''}, where we insert the appropriate class name for each new concept. Note that we employ an LLM only for the sake of convenience (e.g. captioning all 1000 ImageNet classes), but this is not a requirement. For medical data, we actually employ a mix of hand-crafted captions based on Radiopaedia~\citep{radiopaedia} augmented with some elements from ChatGPT-4~\citep{chatgpt}. All captions can be found in the supplementary material.

\paragraph{Inversion}
We run inversion for 5k steps, using a cosine learning rate annealing schedule. For the regularization term, we use the default value $\alpha=0.005$~\citep{kazemi2024we}. The augmentation we employ is composed of random affine transformations (rotation comprised between -30 and +30 degrees, a translation of 10\%, and a scaling comprised between 70\% and 100\% of the image size), with a probability of 0.5. An example of inverted images can be found in Fig.~\ref{fig:inversion-example-full}. For each concept, we generate ten inverted samples. %

\paragraph{Fine-tuning}
Fine-tuning is performed using the InfoNCE loss to achieve alignment between the inverted images and the textual descriptions. We only fine-tune the visual encoder, while keeping the text encoder frozen. The motivation is that we wish to align features extracted from the visual encoder to those extracted from the text encoder. For most experiments, we perform a quick fine-tuning consisting of only one single epoch, with small learning rates between $10^{-6}$ and $10^{-4}$. For CLIP-based models, we generally employ a weight decay of 0.2 as in~\citep{radford2021learning}. More details are provided in the description of each experiment.

\subsection{Datasets}

We employ a variety of datasets in different domains and for different downstream tasks. Here we provide a complete list, divided by task. Note that we do not use any training data from these datasets, as we only use them for testing. All improvements come from the textual description.

\paragraph{Natural images classification} ~\\
\noindent \emph{1.) RareConcepts} is a collection of images of rare concepts gathered from the web. We release the dataset as part of this work. In our experiments, we focus on concepts that are relatively unknown to different large multi-modal architectures: Moongate, Gyroscope and Tonometer, together with four additional fine-grained or deformable categories and geometric patterns (fabric patterns: Bengal Strip, Madras, Floral; parquet pattern: Chantilly). For each concept, we collect 10 images. \\ %
\noindent \emph{2.) ImageNet-1k}~\citep{deng2009imagenet} is a large-scale benchmark for visual recognition, with 1000 classes and 3.2M natural images.
\paragraph{Medical images classification} ~\\
\noindent \emph{3.) CheXpert-2x500c}~\citep{huang2021gloria} is a dataset of Chest X-Rays obtained from the large-scale CheXpert dataset~\citep{irvin2019chexpert} by considering 200 examples for the classes Atelectasis, Cardiomegaly, Edema, Consolidation, and Pleural Effusion.\\
\noindent \emph{4.) JSRT}~\citep{shiraishi2000development} is a Chest X-Ray dataset containing 154 conventional chest radiographs with a lung nodule of different types (malignant and benign nodules).
\paragraph{Medical images segmentation}~\\
\noindent \emph{5.) UnitoChest}~\citep{chaudhry2022unitochest} is a collection of 306,440 chest CT slices coupled with nodules segmentation masks. We consider slices where nodules are present, for a total of 4179 images.\\ 
\noindent \emph{6.) UDIAT}~\citep{yap2017automated} is a dataset of breast masses in ultrasound images, containing 110 benign and 54 malignant cases.\\
\noindent \emph{7.) SIIM Pneumothorax}~\citep{zawacki2019siim} is a Chest X-ray dataset for pneumothorax segmentation, released as a challenge in 2019. We consider a total of 500 images.\\
\noindent \emph{8.) BraTS23 Glioma}~\citep{adewole2023brain} is a brain MRI dataset of adult patients with brain gliomas. We consider all slices where a tumor is present for a total of 14,746 images.
\paragraph{Image-Text retrieval and captioning} ~\\
\noindent \emph{9.) Flickr30k} \citep{young-etal-2014-image} is a dataset of 31,783 images from Flickr, each one associated with 5 captions provided by human annotators. For our experiments, we used Karpathy's test split \citep{karpathy2015deep}, which contains 1000 images and 5000 captions.\\
\noindent \emph{10.) MSCOCO}~\citep{lin2014microsoft} is a large-scale dataset of more than 330k images with textual captions. We use Karpathy's test split~\citep{karpathy2015deep}, containing 5000 images.

\subsection{Controlled Experiments}

\subsubsection{Rare Concepts (CLIP and ViLT)}
\label{sec:app-setup-clip-rare-concepts}

We train using the Adam optimizer, with a batch size of 4, a weight decay of 0.2, and learning rates between 1e-5 and 5e-5 as reported in the table in the main text. We train using 10 inverted images for each concept. The captions used for inversion can be found in Tab.~\ref{tab:captions-rare-concepts}. 

\paragraph{Details about image inversion for ViLT}
For ViLT we use a slightly different approach, in order to accommodate the different architecture. To run inversion, we start from a pair of input $<x_t, \hat{x}^*_v \sim N(0;1)>$ composed by the textual caption and random noise. We then optimize $\hat{x}^*_v$ by optimizing the image-text matching (ITM) score computed on the ITM head of ViLT~\citep{kim2021vilt}. This head outputs two values: one indicating no match and the other indicating a match. To optimize this, we use the cross-entropy loss during inversion, aiming to maximize the output corresponding to a match while minimizing the output for no match.  The rest of the setup is the same as CLIP. Furthermore, we disabled the random affine augmentation, as it produced noisy inverted images. Additionally, we use a weight decay value of 0.01, which is consistent with the one used by the authors of ViLT. 
The captions used for inversion with ViLT can be found in Tab.~\ref{tab:vilt-concept-descriptions}.

\subsubsection{KT on Medical Images (MedCLIP)}
For MedCLIP, we use the same setup as CLIP on rare concepts, see Sec.~\ref{sec:app-setup-clip-rare-concepts}. Namely, we employ Adam with a batch size of 4 and a weight decay of 0.2, using 10 inverted images for each concept. The descriptions used for inversion with MedCLIP can be found in Tab.~\ref{tab:captions-jsrt}. 

\subsubsection{CLIP on medical images (out of domain KT)}
For ViT-B/32, we use a learning rate of 5e-5 with a batch size of 8, and we train for 5 epochs; for ViT-L/14 we use a learning rate of 1e-5, a batch size of 4, and we train for 2 epochs. The captions used for inversion are reported in Tab.~\ref{tab:cations-chexpert-5x200c}.

\subsection{Real-world experiments}

\subsubsection{Captioning (CoCa)}

In these experiments, we deal with two types of captions: the first is the \emph{concept caption}, that we use for inversion and fine-tuning with InfoNCE as in all other experiments (listed in Sec.~\ref{sec:all-captions}), the second is the \emph{target caption} that we use to fine-tune the autoregressive captioning decoder of CoCa with $\mathcal{L}_{cap}$. 
 
\paragraph{Captioning Loss}

When fine-tuning on inverted images, we apply an autoregressive captioning loss, as defined in~\citep{yu2022coca}:

\begin{equation}
    \mathcal{L}_{cap} = - \sum_{t=1}^T \log P_\theta(y_t | y_{<t},x)
\end{equation}

\noindent which aims at predicting the next token $y_t$ given the previous tokens $y_{<t}$ and the image $x$. The final objective function that we optimize is the combination of the InfoNCE loss and the captioning loss:

\begin{equation}
    \mathcal{L} = \lambda_1 \mathcal{L}_{CLIP} + \lambda_2 \mathcal{L}_{cap}
\end{equation}

\noindent where $\lambda_1, \lambda_2 \geq 0$. In our fine-tuning, we use $\lambda_1=1$ and $\lambda_2 = 0.1$.

\paragraph{Target captions template}

We use a set of 26 different templates as target captions during fine-tuning. At each optimization step, we select a random template for each sample in the following manner:

\begin{lstlisting}[language=Python]
TEMPLATES = (
 lambda c: f'a bad photo of a {c}.',
 lambda c: f'a low resolution photo of the {c}.',
 lambda c: f'a rendering of a {c}.',
 lambda c: f'a bad photo of the {c}.',
 lambda c: f'a cropped photo of the {c}.',
 lambda c: f'a photo of a hard to see {c}.',
 lambda c: f'a bright photo of a {c}.',
 lambda c: f'a photo of a clean {c}.',
 lambda c: f'a photo of a dirty {c}.',
 lambda c: f'a dark photo of the {c}.',
 lambda c: f'a photo of my {c}.',
 lambda c: f'a bright photo of the {c}.',
 lambda c: f'a cropped photo of a {c}.',
 lambda c: f'a photo of the {c}.',
 lambda c: f'a good photo of the {c}.',
 lambda c: f'a rendering of the {c}.',
 lambda c: f'a photo of one {c}.',
 lambda c: f'a close-up photo of the {c}.',
 lambda c: f'a photo of a {c}.',
 lambda c: f'a low resolution photo of a {c}.',
 lambda c: f'a photo of a large {c}.',
 lambda c: f'itap of the {c}.',
 lambda c: f'a jpeg corrupted photo of the {c}.',
 lambda c: f'a good photo of a {c}.',
 lambda c: f'itap of a {c}.',
 lambda c: f'a photo of the large {c}.',
)

template_idx = torch.randint(
    0, len(TEMPLATES), (1,)
).item()
template = TEMPLATES[template_idx]
return tokenize(template(class_name))
\end{lstlisting}

These templates are inspired by OpenAI prompt ensembling for zero-shot classifiers\footnote{\url{https://github.com/mlfoundations/open_clip/blob/main/src/open_clip/zero_shot_metadata.py}}. We use these captions although they are not in the exact style of MSCOCO, as we do not want to leverage information from MSCOCO besides the concept classes. Target captions crafted specifically for MSCOCO might further improve the results.

\section{Additional Results}

\subsection{Controlled Experiments}

\subsubsection{Rare Concepts (CLIP and ViLT)}

\begin{figure*}
    \includegraphics[width=\textwidth]{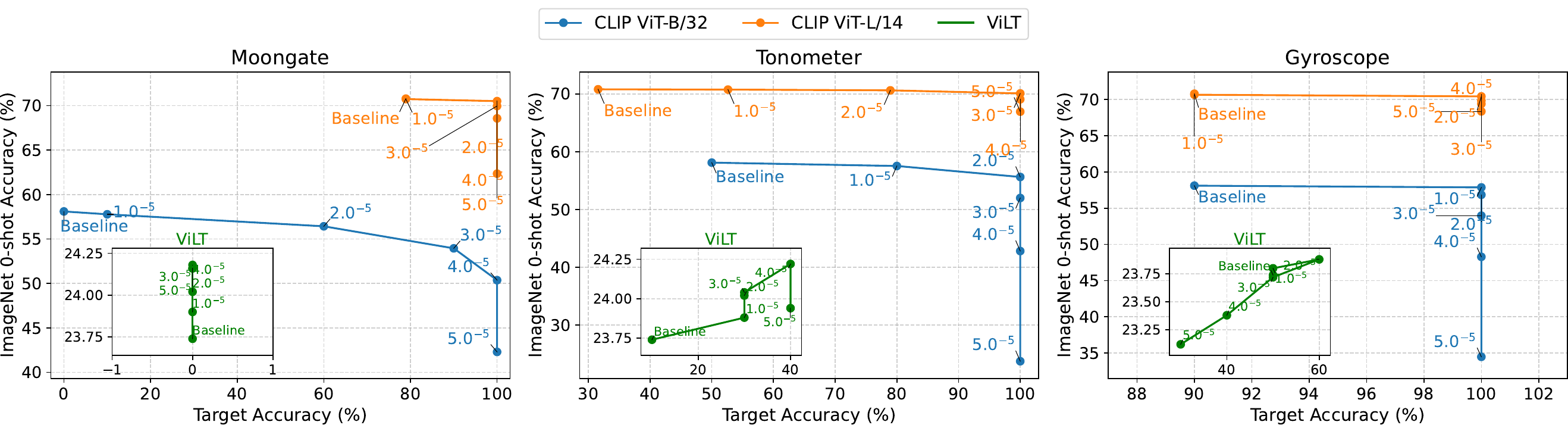}
    \caption{\textbf{Knowledge Transfer (KT) on novel and rare concepts} (CLIP and ViLT) across different learning rates. \e{In most instances, we achieve improvement (even notable) in the target accuracy on the novel concept, preserving original knowledge (measured as accuracy on ImageNet). We also observe that on ViLT the accuracy on ImageNet generally improves when performing KT.}} %
    \label{fig:kt-results-lr}
\end{figure*}

Here we report results across differen learning rate values, in Fig.~\ref{fig:kt-results-lr}. Results in numerical forms can be found in Tab.\ref{tab:kt-results-lr}.

\begin{table*}
    \centering
    \resizebox{0.9\linewidth}{!}{%
    \begin{tabular}{l l l c c c c c c c}
        \toprule
         & & & & \multicolumn{5}{c}{\textbf{Learning Rate}} \\
         \cmidrule(l){5-9}
         \textbf{Model} & \textbf{Concept} & & \textbf{Baseline} & \textbf{1e-5} & \textbf{2e-5} & \textbf{3e-5} & \textbf{4e-5} & \textbf{5e-5}  \\
         \midrule
         CLIP ViT-B/32~\citep{radford2021learning} & Moongate & Target Acc. & 0\% & 10\% & \textbf{60\%} & 90\% & 100\% & 100\% \\
         & & ImageNet 0-shot & 58.10\% & 57.78\% & \textbf{56.43\%} & 53.95\% & 50.37\% & 42.30\% \\
         \cmidrule(l){2-9}
         & Tonometer & Target Acc. & 50\% & \textbf{80\%} & 80\% & 100\% & 100\% & 100\% \\
         & & ImageNet 0-shot & 58.10\% & \textbf{57.52\%} & 55.62\% & 51.98\% & 42.80\% & 23.73\% \\
         \cmidrule(l){2-9}
         & Gyroscope & Target Acc. & 90\% & \textbf{100\%} & 100\% & 100\% & 100\% & 100\% \\
         & & ImageNet 0-shot & 58.10\% & \textbf{57.86\%} & 56.84\% & 53.96\% & 48.28\% & 34.48\% \\
         \midrule
         \midrule
         CLIP ViT-L/14~\citep{radford2021learning} & Moongate & Target Acc. & 78.95\% & 78.95\% & \textbf{100\%} & 100\% & 100\% & 100\% \\
         & & ImageNet 0-shot & 70.79\% & 70.74\% & \textbf{70.51\%} & 69.96\% & 68.57\% & 62.35\% \\
         \cmidrule(l){2-9}
         & Tonometer & Target Acc. & 31.58\% & 52.63\% & \textbf{78.95\%} & 100\% & 100\% & 100\% \\
         & & ImageNet 0-shot & 70.79\% & 70.74\% & \textbf{70.61\%} & 70.08\% & 69.06\% & 66.92\% \\
         \cmidrule(l){2-9}
         & Gyroscope & Target Acc. & 90\% & 90\% & \textbf{100\%} & 100\% & 100\% & 100\% \\
         & & ImageNet 0-shot & 70.79\% & 70.65\% & \textbf{70.42\%} & 69.84\% & 69.39\% & 68.35\% \\
         \midrule
         \midrule
         ViLT~\citep{kim2021vilt} & Moongate & Target Acc. & 0\% & 0\% & 0\% & 0\% & \textbf{0\%} & 0\% \\
         & & ImageNet* 0-shot & 23.74\% & 23.90\% & 24.02\% & 24.16\% & \textbf{24.18\%} & 24.16\% \\
         \cmidrule(l){2-9}
         & Tonometer & Target Acc. & 10\% & 30\% & 30\% & 30\% & \textbf{40\%} & 40\% \\
         & & ImageNet* 0-shot & 23.74\% & 23.88\% & 24.02\% & 24.04\% & \textbf{24.22\%} & 23.94\% \\
         \cmidrule(l){2-9}
         & Gyroscope & Target Acc. & 50\% & \textbf{60\%} & 50\% & 50\% & 40\% & 30\% \\
         & & ImageNet* 0-shot & 23.74\% & \textbf{23.80\%} & 23.88\% & 23.72\% & 23.38\% & 23.12\% \\
         \bottomrule
         & 
    \end{tabular}
    }
    \caption{\textbf{Knowledge Transfer on novel and rare concepts} (CLIP and ViLT) in terms of accuracy. * for VilT, we employ ImageNet-100~\citep{imagenet100} due to the computational requirements of evaluating every possible image-caption pair for zero-shot classification.}
    \label{tab:kt-results-lr}
\end{table*}

\subsubsection{KT on Fine-grained and deformable rare-concepts}

Tab.~\ref{tab:textures}, we perform additional experiments on fine-grained categories, including both deformable (fabric) and non-deformable (parquet) patterns. 

\begin{table}[!h]
  \resizebox{\linewidth}{!}{%
    \begin{tabular}{l c c c| c}
      \toprule
      & \multicolumn{3}{c|}{\textbf{Fabric}} & \textbf{Parquet} \\
      \textbf{Model} & \textbf{Bengal Stripe} & \textbf{Floral} & \textbf{Madras} & \textbf{Chantilly} \\
      \midrule
      CLIP ViT-B/32  & 0\% & 10\% & 0\% & 20\% \\
      \rowcolor{Apricot} \textbf{CLIP ViT-B/32 + KT} (ours)  & \textbf{20\%} & \textbf{30\%} & \textbf{10\%} & \textbf{70\%} \\
      ImageNet (KT) & 57.08\% & 57.63\% & 57.33\% & 56.12\% \\ 
      \midrule
      CLIP ViT-L/14 & 70\% & 60\% & 80\% & 0\%\\
      \rowcolor{Apricot} \textbf{CLIP ViT-L/14 + KT} (ours) & \textbf{90\%} & \textbf{80\%} & \textbf{90\%} & \textbf{10\%} \\
      ImageNet (KT) & 70.82\% & 70.73\% & 70.79\% & 69.90\% \\
      \bottomrule
    \end{tabular}%
  }
  \vspace{2pt}
  \caption{\textbf{KT applied to fine-grained and deformable categories} (fabric and parquet patterns), expanding the Rare Concepts dataset.}
  \label{tab:textures}
\end{table}

\subsubsection{KT on medical images (MedCLIP)}
The full results on JSRT with MedCLIP can be found in Tab.~\ref{tab:kt-results-medclip-lr-full}.

\begin{table*}
    \centering
    \resizebox{0.8\linewidth}{!}{%
    \begin{tabular}{l l c c c c c c}
        \toprule
         & & & \multicolumn{5}{c}{\textbf{Learning Rate (multiplier)}} \\
         \cmidrule{4-8}
         \textbf{Concept} & & \textbf{Baseline} & \textbf{$\times$1} & \textbf{$\times$2} & \textbf{$\times$3} & \textbf{$\times$4} & \textbf{$\times$5}  \\
         \midrule
         Benign Nodule & Target Acc. (base lr 1e-5) & 54.55\% & 54.55\% & \textbf{54.55\%} & 54.55\% & 54.55\% & 54.55\% \\
         & CheXpert-5x200c 0-shot & 62.10\% & 61.80\% & \textbf{62.30\%} & 62.10\% & 62\% & 62.20\% \\
         \midrule
         Lung Cancer & Target Acc. (base lr 1e-4) & 83.93\% & 87.50\%  & \textbf{92.86\%} & 94.64\% & 92.86\%  & 92.86\% \\
         & CheXpert-5x200c 0-shot & 62.10\% & 62.20\% & \textbf{61.50\%} & 53.70\% & 48.20\% & 44.50\%\\
         \bottomrule
    \end{tabular}
    }
    \caption{\textbf{Knowledge Transfer on MedCLIP on the JSRT dataset} (accuracy). Full results across learning rates.}
    \label{tab:kt-results-medclip-lr-full}
\end{table*}

\begin{table*}
    \centering
    \resizebox{\linewidth}{!}{
    \begin{tabular}{l c c c c c c c c c c c c}
    \toprule
    &  \multicolumn{3}{c}{\textbf{Lung Nodules$^\dagger$}} & \multicolumn{3}{c}{\textbf{Lung Pneumothorax}$^\dagger$} & \multicolumn{3}{c}{\textbf{Breast Ultrasound}} & \multicolumn{3}{c}{\textbf{Brain MRI}}  \\
    \cmidrule(l){2-4} \cmidrule(l){5-7} \cmidrule(l){8-10} \cmidrule(l){11-13} 
    \textbf{Model} & \textbf{DSC} & \textbf{NSD} & \textbf{IoU} & \textbf{DSC} & \textbf{NSD} & \textbf{IoU} & \textbf{DSC} & \textbf{NSD} & \textbf{IoU} & \textbf{DSC} & \textbf{NSD} & \textbf{IoU} \\
    \midrule
    MedCLIP-SAMv2 & \textbf{14.83}\% & 17.30\% & 8.64\% & 6.30\% & 7.61\% & 3.75\% 
    & 56.25\% & 59.44\% & 47.81\% & 17.20\% & 20.97\% & 12.05\% \\
    Transf. (1e-5) & 13.95\% & 17.45\% & 8.75\% & 6.28\% & 7.59\% & 3.77\% & \textbf{58.23}\% & \textbf{61.56}\% & \textbf{49.52}\%  & 15.90\% & 19.36\% & 11.10\% \\
    Transf. (2e-5) & 14.10\% & 17.65\% & 8.83\% & \textbf{6.41}\% & \textbf{7.76}\% & \textbf{3.83}\% & 54.36\% & 57.30\% & 46.30\% & \textbf{18.13}\% & \textbf{22.26}\% & \textbf{12.62}\% \\
    Transf. (3e-5) & 14.10\% & 17.65\% & 8.85\% & 6.25\% & 7.55\% & 3.73\% & 55.70\% & 59.00\% & 47.49\% & 15.47\% & 18.85\% & 10.78\% \\
    Transf. (4e-5) & 14.25\% & 17.85\% & 8.94\% & 6.24\% & 7.57\% & 3.71\% & 53.86\% & 56.82\% & 45.61\% & 15.26\% & 18.63\% & 10.62\% \\
    Transf. (5e-5) & 14.20\% & 17.78\% & 8.92\% & 6.20\% & 7.51\% & 3.70\% & 54.90\% & 57.97\% & 46.09\% & 16.22\% & 19.81\% & 11.34\% \\
    Transf. (1e-4) & 14.35\% & \textbf{18.03}\% & \textbf{9.04}\% & 6.02\% & 7.29\% & 3.59\% & - & - & - & - & - & -\\
    Transf. (2e-4) & 10.74\% & 13.64\% & 6.66\% & 4.71\% & 5.54\% & 2.86\% & - & - & - & - & - & -\\
    \bottomrule
    \end{tabular}
    }
    \caption{Full results on zero-shot segmentation with MedCLIP-SAMv2.}
    \label{tab:medclip-samv2-results-full}
\end{table*}

\subsection{Real-world Experiments}

\subsubsection{Segmentation}

\begin{figure*}
    \centering
    \includegraphics[width=0.15\linewidth]{img/segmentation/breast/0_image.png}
    \includegraphics[width=0.15\linewidth]{img/segmentation/breast/0_gt_mask.png}
    \includegraphics[width=0.15\linewidth]{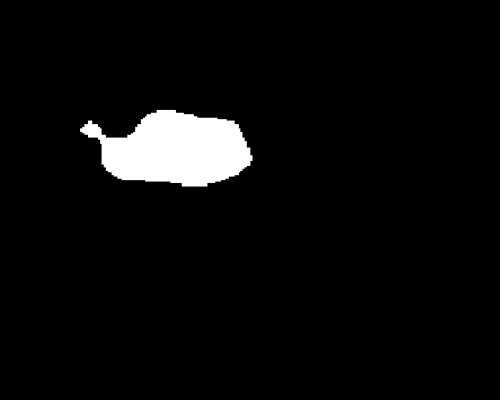}
    \includegraphics[width=0.15\linewidth]{img/segmentation/breast/0_baseline_pred_mask.png}
    \includegraphics[width=0.15\linewidth]{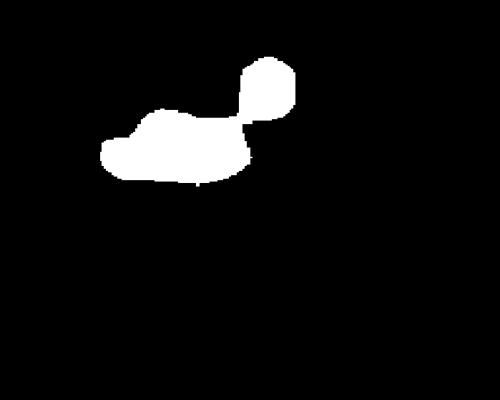}
    \includegraphics[width=0.15\linewidth]{img/segmentation/breast/0_pred_mask.png} \\
    \includegraphics[width=0.15\linewidth]{img/segmentation/breast/1_image.png}
    \includegraphics[width=0.15\linewidth]{img/segmentation/breast/1_gt_mask.png}
    \includegraphics[width=0.15\linewidth]{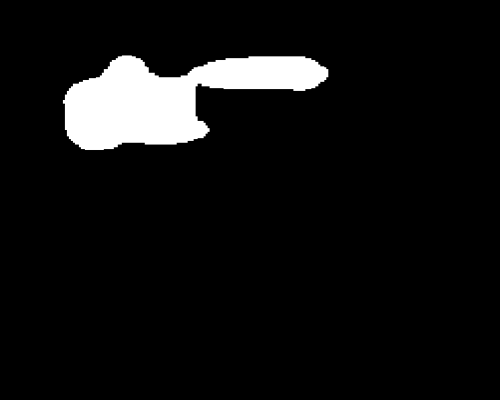}
    \includegraphics[width=0.15\linewidth]{img/segmentation/breast/1_baseline_pred_mask.png}
    \includegraphics[width=0.15\linewidth]{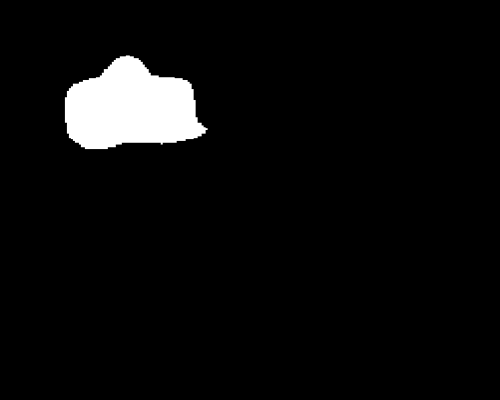}
    \includegraphics[width=0.15\linewidth]{img/segmentation/breast/1_pred_mask.png}
    \includegraphics[width=0.15\linewidth]{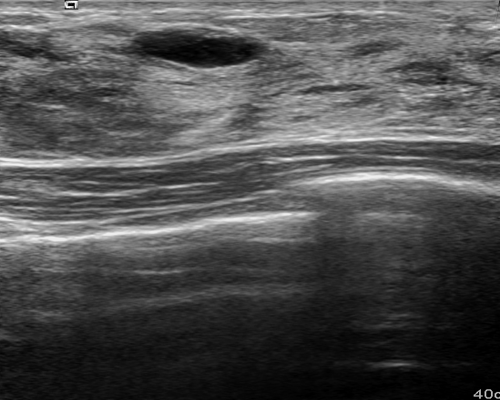}
    \includegraphics[width=0.15\linewidth]{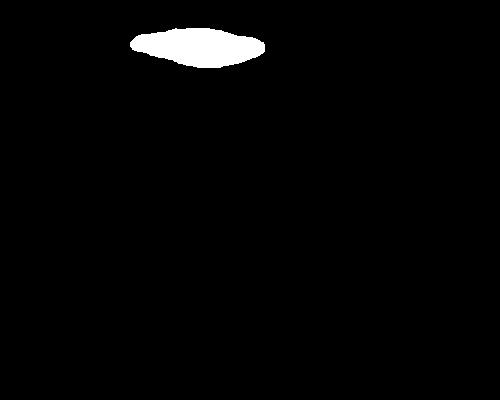}
    \includegraphics[width=0.15\linewidth]{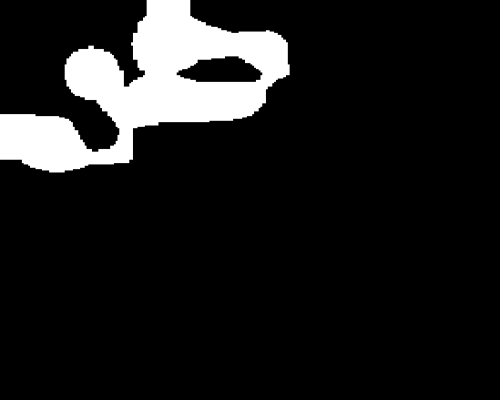}
    \includegraphics[width=0.15\linewidth]{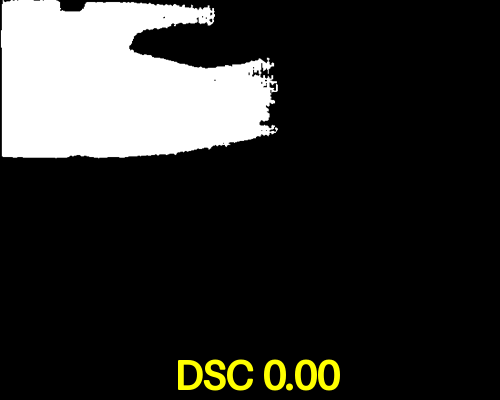}
    \includegraphics[width=0.15\linewidth]{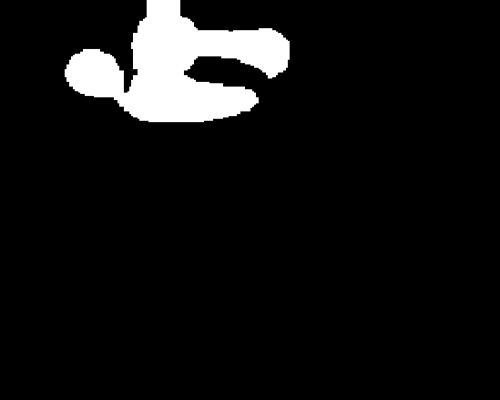}
    \includegraphics[width=0.15\linewidth]{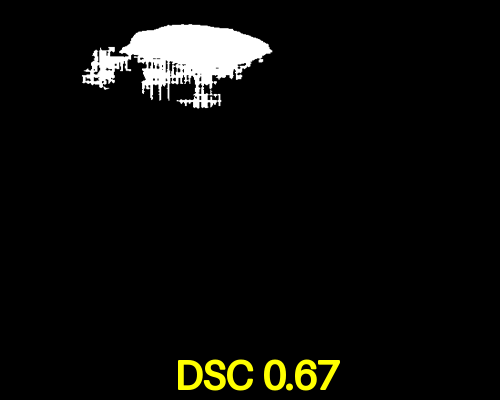}
    \includegraphics[width=0.15\linewidth]{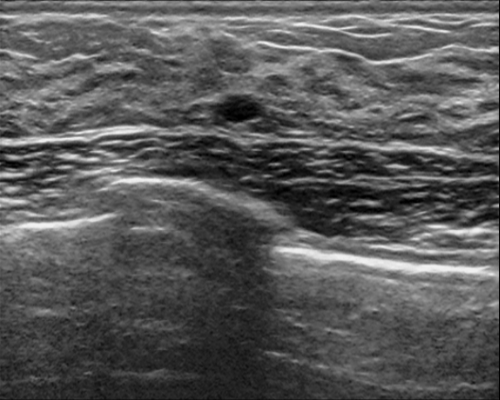}
    \includegraphics[width=0.15\linewidth]{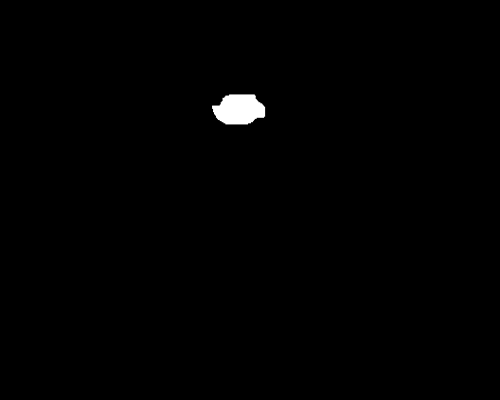}
    \includegraphics[width=0.15\linewidth]{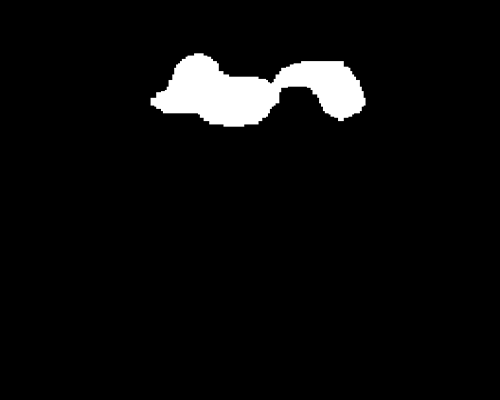}
    \includegraphics[width=0.15\linewidth]{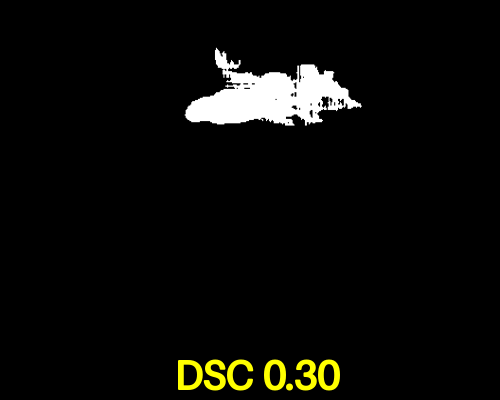}
    \includegraphics[width=0.15\linewidth]{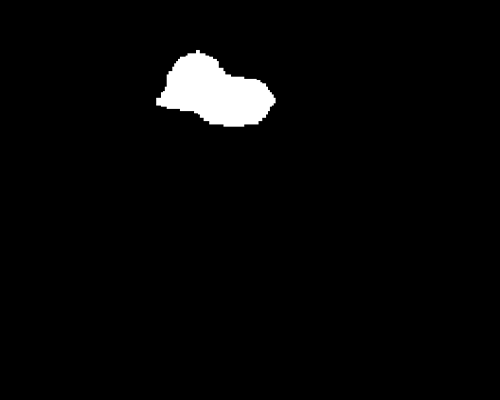}
    \includegraphics[width=0.15\linewidth]{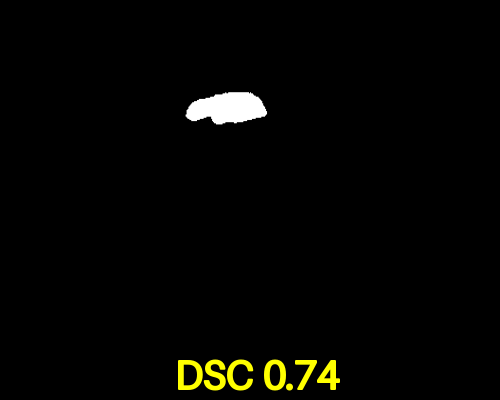}
    \begin{subfigure}[t]{0.15\linewidth}
        \includegraphics[width=\linewidth]{img/segmentation/breast/4_image.png}
        \caption{Image}
    \end{subfigure}
    \begin{subfigure}[t]{0.15\linewidth}
        \includegraphics[width=\linewidth]{img/segmentation/breast/4_gt_mask.png}
        \caption{Ground Truth}
    \end{subfigure}
     \begin{subfigure}[t]{0.15\linewidth}
        \includegraphics[width=\linewidth]{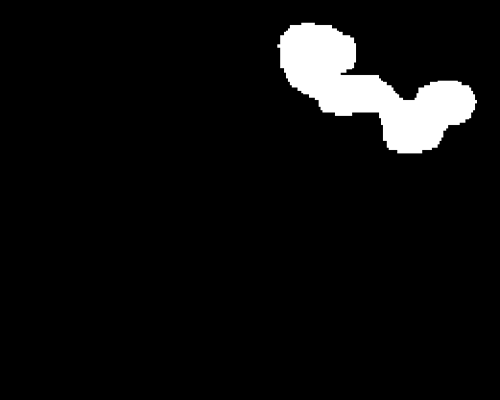}
        \caption{M2IB Map\\(Baseline)}
    \end{subfigure}
    \begin{subfigure}[t]{0.15\linewidth}
        \includegraphics[width=\linewidth]{img/segmentation/breast/4_baseline_pred_mask.png}
        \caption{Final Segmentation (Baseline)}
    \end{subfigure}
     \begin{subfigure}[t]{0.15\linewidth}
        \includegraphics[width=\linewidth]{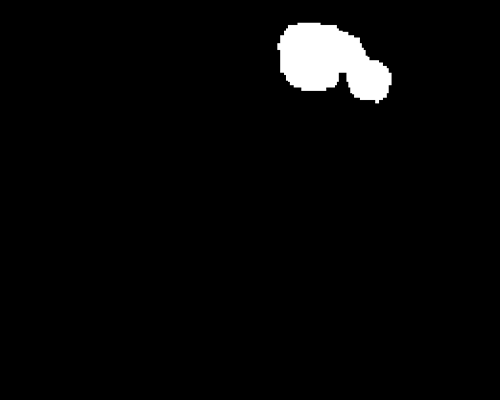}
        \caption{M2IB Map\\(Knowledge Transfer)}
    \end{subfigure}
    \begin{subfigure}[t]{0.15\linewidth}
        \includegraphics[width=\linewidth]{img/segmentation/breast/4_pred_mask.png}
        \caption{Final Segmentation (Knowledge Transfer)}
    \end{subfigure}
    \caption{Qualitative evaluation of knowledge transfer on breast tumor segmentation (UDIAT dataset). We report the top ten most illustrative examples in which knowledge transfer improved segmentation, in terms of DSC.}
    \label{fig:segmentation-example-breast}
\end{figure*}

\begin{figure*}
    \centering
    \includegraphics[width=0.15\linewidth]{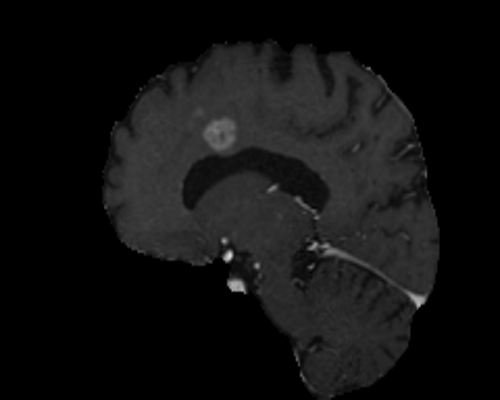}
    \includegraphics[width=0.15\linewidth]{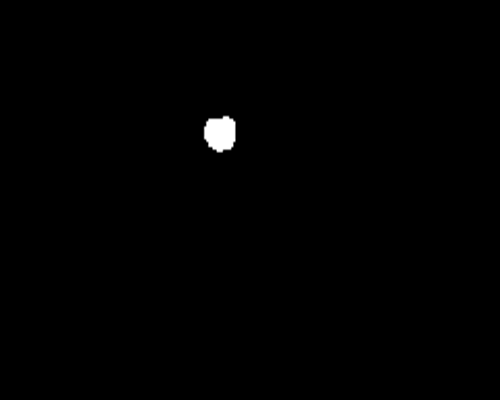}
    \includegraphics[width=0.15\linewidth]{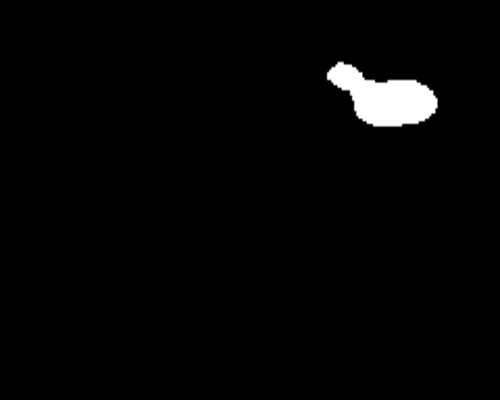}
    \includegraphics[width=0.15\linewidth]{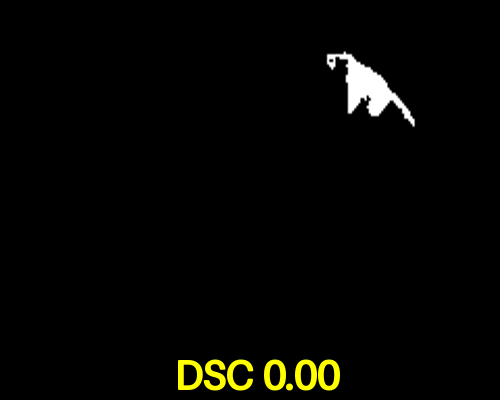}
    \includegraphics[width=0.15\linewidth]{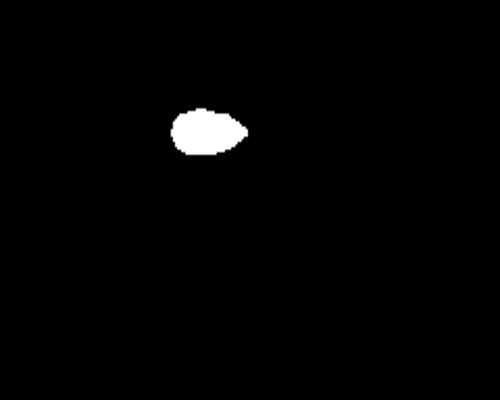}
    \includegraphics[width=0.15\linewidth]{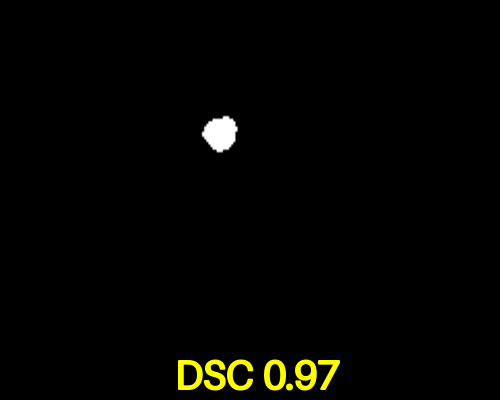} \\
    \includegraphics[width=0.15\linewidth]{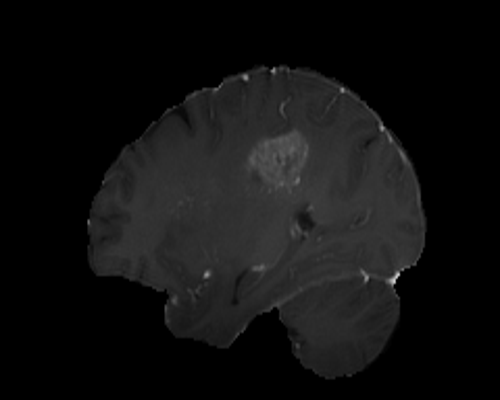}
    \includegraphics[width=0.15\linewidth]{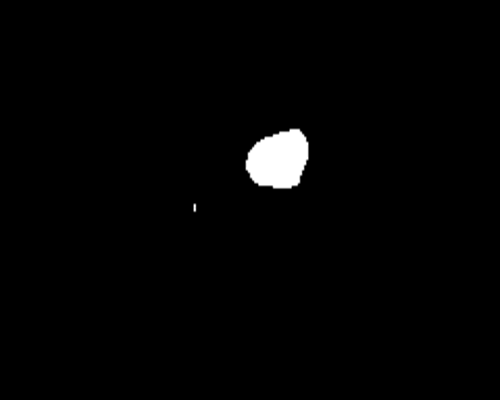}
    \includegraphics[width=0.15\linewidth]{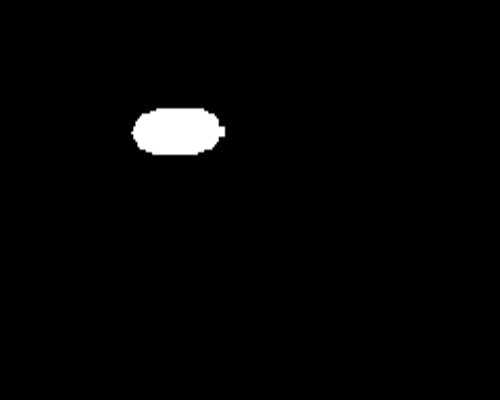}
    \includegraphics[width=0.15\linewidth]{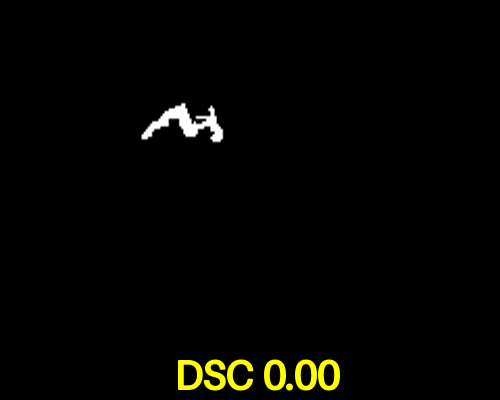}
    \includegraphics[width=0.15\linewidth]{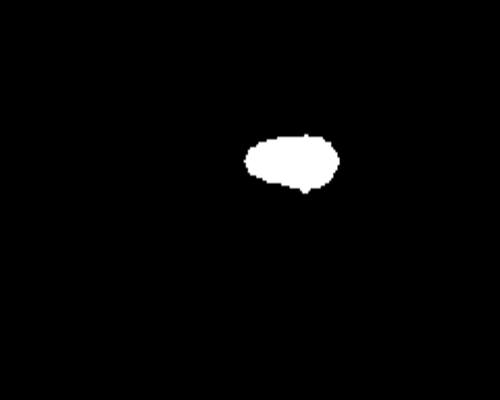}
    \includegraphics[width=0.15\linewidth]{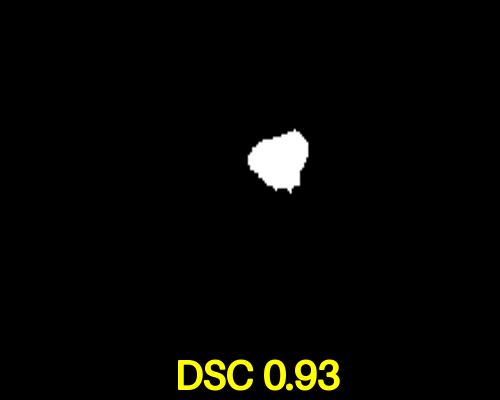}
    \includegraphics[width=0.15\linewidth]{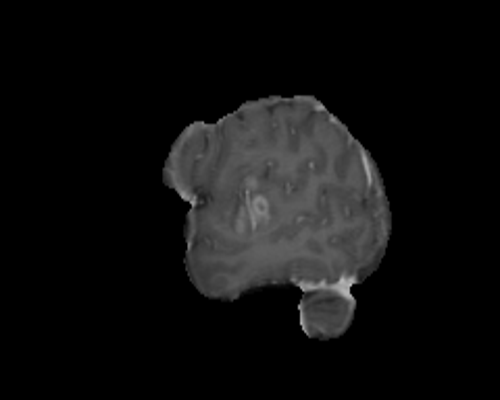}
    \includegraphics[width=0.15\linewidth]{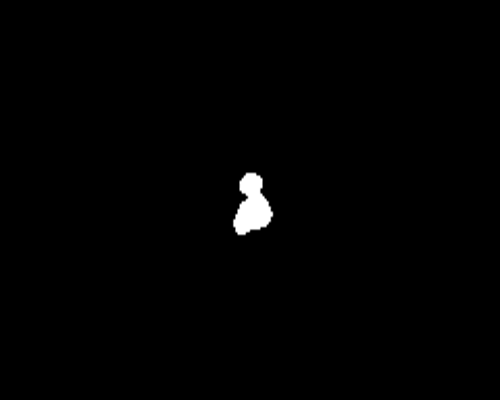}
    \includegraphics[width=0.15\linewidth]{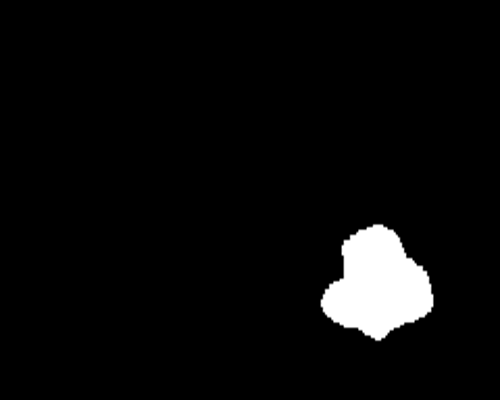}
    \includegraphics[width=0.15\linewidth]{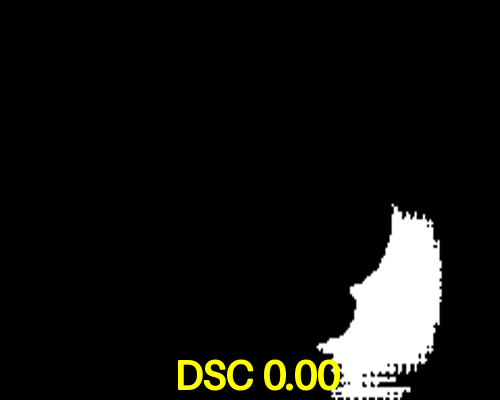}
    \includegraphics[width=0.15\linewidth]{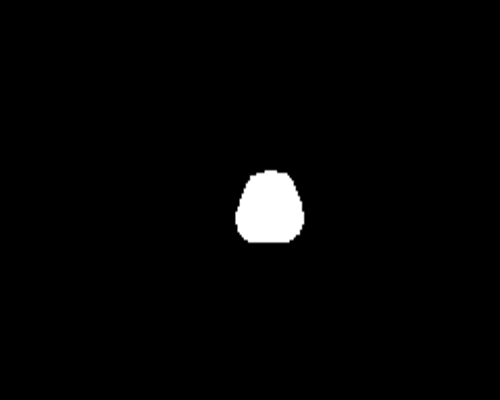}
    \includegraphics[width=0.15\linewidth]{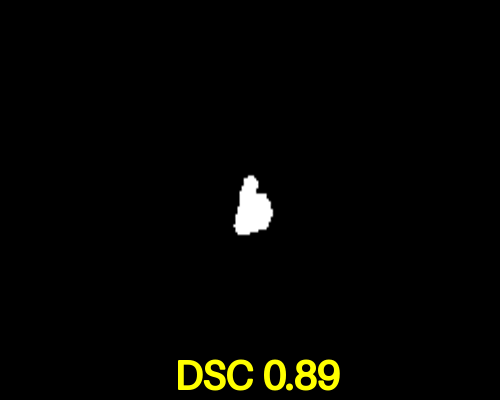}
    \includegraphics[width=0.15\linewidth]{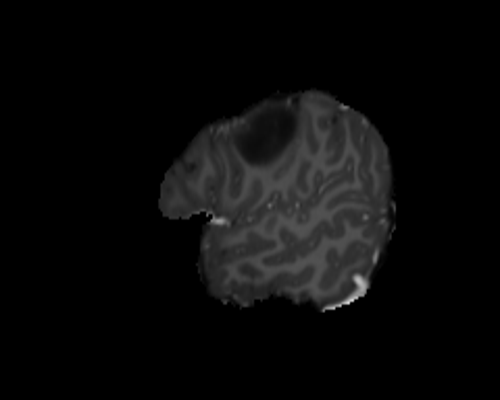}
    \includegraphics[width=0.15\linewidth]{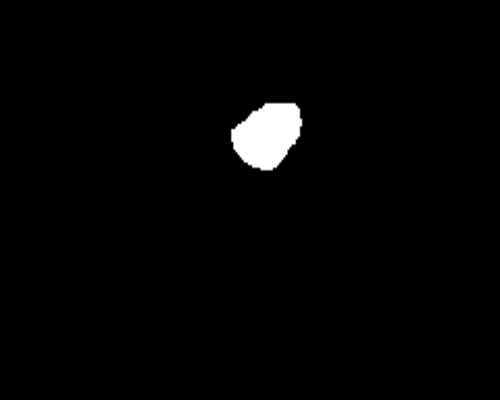}
    \includegraphics[width=0.15\linewidth]{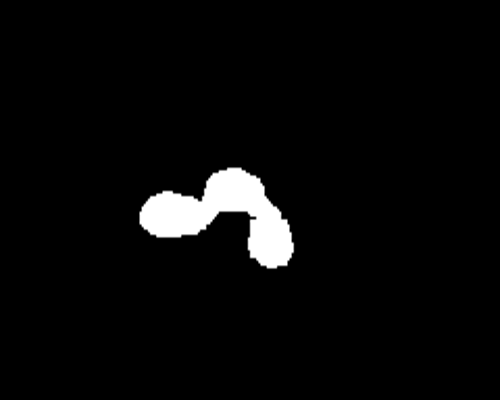}
    \includegraphics[width=0.15\linewidth]{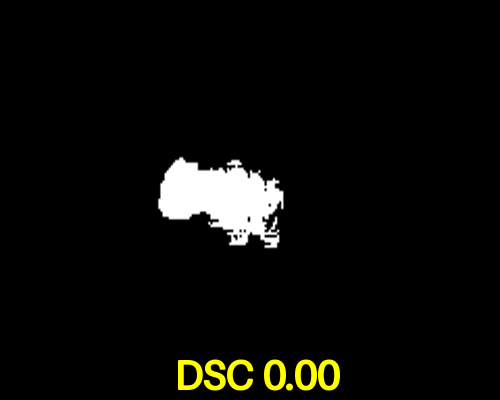}
    \includegraphics[width=0.15\linewidth]{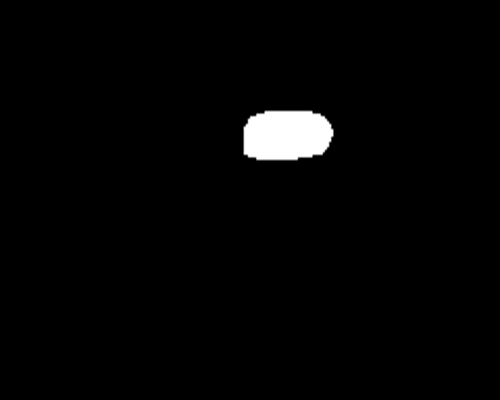}
    \includegraphics[width=0.15\linewidth]{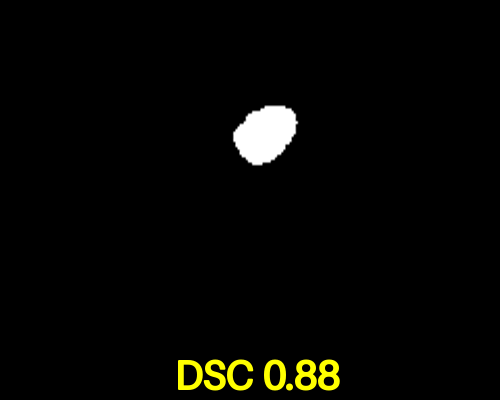}
    \begin{subfigure}[t]{0.15\linewidth}
        \includegraphics[width=\linewidth]{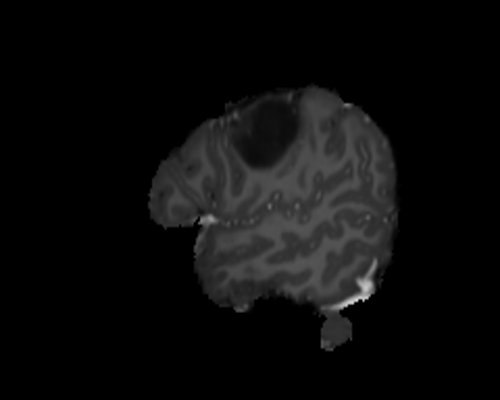}
        \caption{Image}
    \end{subfigure}
    \begin{subfigure}[t]{0.15\linewidth}
        \includegraphics[width=\linewidth]{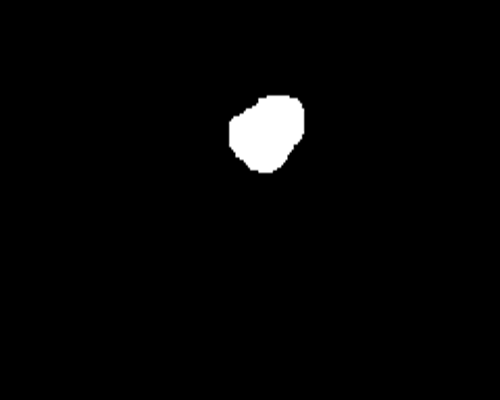}
        \caption{Ground Truth}
    \end{subfigure}
     \begin{subfigure}[t]{0.15\linewidth}
        \includegraphics[width=\linewidth]{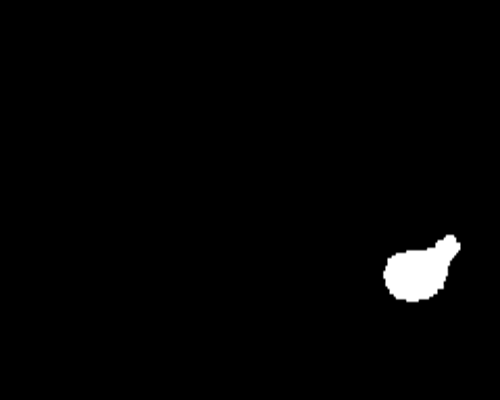}
        \caption{M2IB Map\\(Baseline)}
    \end{subfigure}
    \begin{subfigure}[t]{0.15\linewidth}
        \includegraphics[width=\linewidth]{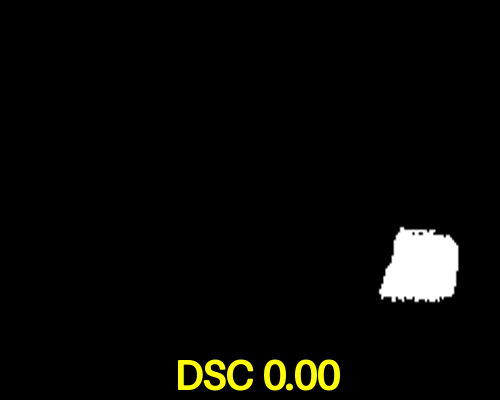}
        \caption{Final Segmentation (Baseline)}
    \end{subfigure}
     \begin{subfigure}[t]{0.15\linewidth}
        \includegraphics[width=\linewidth]{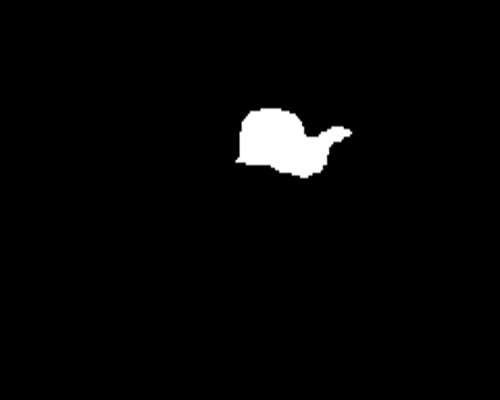}
        \caption{M2IB Map\\(Knowledge Transfer)}
    \end{subfigure}
    \begin{subfigure}[t]{0.15\linewidth}
        \includegraphics[width=\linewidth]{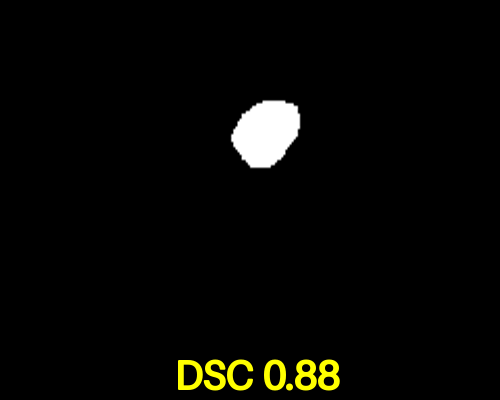}
        \caption{Final Segmentation (Knowledge Transfer)}
    \end{subfigure}
    \caption{Qualitative evaluation of knowledge transfer on brain tumor segmentation (BraTS 2023 glioma dataset). We report the top ten most illustrative examples in which knowledge transfer improved segmentation, in terms of DSC.}
    \label{fig:segmentation-example-brain}
\end{figure*}

Results of knowledge transfer on MedCLIP-SAMv2 with different values of learning rate are shown in Tab.~\ref{tab:medclip-samv2-results-full}.
We report illustrative examples of the improvements achieved by knowledge transfer in Fig.~\ref{fig:segmentation-example-breast} and Fig.~\ref{fig:segmentation-example-brain}. The captions used for inversion for segmentation can be found in Tab.~\ref{tab:captions-segmentation}.

\paragraph{Differences in downstream tasks}

As said in the main text, lung nodules and pneumothorax segmentations are novel tasks on which MedCLIP-SAMv2 was not pre-trained. Regarding brain tumors, we employ the BraTS 2023 glioma dataset, which contains brain gliomas in adult patients. With respect to the original performance reported in~\citep{koleilat2024medclipsamv2} on brain tumors, we notice a significant gap. However, the preprocessing of the images is quite different, as data from BraTS 2023 is more heavily preprocessed (e.g. skull stripping) than in~\citep{koleilat2024medclipsamv2}. We were not able to compare MedCLIP-SAMv2 on the original data, as, at the time of writing, details about the data split are missing.

\subsubsection{Text-image retrieval}

\begin{table*}
\centering
\begin{tabular}{@{}lcccccccc@{}}
\toprule
& & & \multicolumn{6}{c}{\textbf{Flickr30k (1K)}} \\
& & & \multicolumn{3}{c}{\textbf{Text Retrieval}} & \multicolumn{3}{c}{\textbf{Image Retrieval}} \\ 
\cmidrule(l){4-6} \cmidrule(l){7-9} 
\textbf{Model} & LR & Batch Size & \textbf{R@1} & \textbf{R@5} & \textbf{R@10} & \textbf{R@1} & \textbf{R@5} & \textbf{R@10} \\ 
\midrule
ViLT-B/32 (huggingface) & - & - & 73.8\% & 93.5\% & 96.5\% & 57.3\% & 83.9\% & 90.4\% \\
\midrule
ViLT-B/32 & 8e-7 & 32  & 74.5\%          & 93.8\%          & 96.4\%          & 57.7\%          & 84.0\%          & 90.4\%          \\
ViLT-B/32 & 9e-7 & 32  & \textbf{74.6\%} & 93.8\%          & 96.4\%          & \textbf{57.8\%} & 84.0\%          & 90.5\%          \\
ViLT-B/32 & 1e-6 & 16  & 74.4\%          & 93.8\%          & 96.5\%          & 57.7\%          & \textbf{84.1\%} & 90.5\%          \\
ViLT-B/32 & 2e-6 & 128 & \textbf{74.6\%} & 93.7\%          & 96.5\%          & \textbf{57.8\%} & 84.0\%          & 90.5\%          \\
ViLT-B/32 & 3e-6 & 256 & 74.5\%          & \textbf{93.9\%} & 96.5\%          & 57.7\%          & 83.9\%          & 90.5\%          \\
ViLT-B/32 & 4e-6 & 32  & 73.8\%          & 93.6\%          & 96.5\%          & 57.4\%          & 84.0\%          & 90.5\%          \\
ViLT-B/32 & 5e-6 & 256 & 74.5\%          & \textbf{93.9\%} & 96.5\%          & 57.6\%          & 84.0\%          & 90.5\%          \\
ViLT-B/32 & 8e-6 & 32  & 73.2\%          & 93.7\%          & 96.1\%          & 57.4\%          & 83.7\%          & 90.4\%          \\
ViLT-B/32 & 1e-5 & 128 & 74.4\%          & 93.8\%          & \textbf{96.8\%} & 56.8\%          & 83.7\%          & \textbf{90.6\%} \\
ViLT-B/32 & 2e-5 & 32  & 71.8\%          & 93.2\%          & 96.4\%          & 56.7\%          & 83.6\%          & 90.4\%          \\
ViLT-B/32 & 3e-5 & 32  & 70.8\%          & 92.1\%          & 95.7\%          & 56.0\%          & 82.9\%          & 90.2\%          \\
\bottomrule
\end{tabular}%
\caption{Full results for text and image retrieval on Flickr30k with ViLT. The first section reports baseline results, while the second shows the outcome of each tested learning rate and its optimal batch size (chosen among 16, 32, 64, 128, and 256). %
Recall scores at top 1, 5, and 10 are reported.}
\label{tab:flickr30k-eval-full-results}
\end{table*}

In this section, we show the full results for text and image retrieval tasks on Flickr30k with ViLT. Tab.~\ref{tab:flickr30k-eval-full-results} is the extended version of the results in which we report the huggingface's pre-trained baseline, along with the results of the experiments we performed while tuning the learning rate and the batch size. We report the best batch size for each learning rate. As can be seen our method works best with smaller learning rates in this setting. The captions used for inversion (mscoco) can be found in Tab.~\ref{tab:captions-mscoco}.

\subsubsection{Captioning}

We report additional results of captioning, with and without the use of captioning loss in Tab.~\ref{tab:mscoco-captioning-results-all}. Even without captioning loss, we are achieve improvements over both version CoCa (pre-trained on LAION-2B) and CoCa FT (fine-tuned on MSCOCO). By applying $\mathcal{L}_{cap}$ we achieve a further increase in the reported metrics. We showcase some improvements in captioning on the CoCa model pre-trained on LAION-2B in Fig.~\ref{fig:mscoco-captioning-examples}. The concept captions used for inversion can be found in Tab.~\ref{tab:captions-mscoco}.

\begin{table}
    \centering
    \resizebox{\columnwidth}{!}{%
    \begin{tabular}{l c c c c}
    \toprule
    & \multicolumn{4}{c}{\textbf{MSCOCO (5K)}} \\
    \textbf{Model} & \textbf{BLEU@4} &\textbf{METEOR} & \textbf{CIDEr} & \textbf{SPICE} \\ 
    \midrule
    CLIP-ViL~\citep{shen2022how} & 40.2 & 29.7 & 134.2 & 23.8 \\
    BLIP~\citep{li2022blip} & 40.4 & - & 136.7 & - \\
    VinVL~\citep{zhang2021vinvl} & 41.0 & 31.1 & 140.9 & \textbf{25.4} \\
    SimVLM~\citep{wang2022simvlm} & 40.6 & 33.7 & 143.3 & 25.4 \\
    LEMON~\citep{hu2021scaling} & \textbf{41.5} & 30.8 & 139.1 & 24.1 \\
    CoCa~\citep{yu2022coca} (proprietary) & 40.9 & \textbf{33.9} & \textbf{143.6} & 24.7 \\
    \midrule
    CoCa & 6.9 & 12.8 & 31.1 & 9.1 \\
    CoCa (transf. 6e-5) & 13.6 & 18.5 & 47.3 & 13.6 \\
    CoCa$^\dagger$ (transf. 9e-5) & \textbf{17.9} & \textbf{19.4} & \textbf{60.8} & \textbf{13.7} \\
    \midrule 
    CoCa FT  & 34.9 & 29.7 & 123.1 & \textbf{23.5} \\
    CoCa FT (transf. 2e-5) & 35.2 & 29.8 & 123.1 & 23.2 \\
    CoCa FT$^\dagger$ (transf. 5e-6) & \textbf{35.2} & \textbf{29.8} & \textbf{124.0} & 23.3 \\
    
    \bottomrule
    \end{tabular}
    }
    \caption{Image captioning on MSCOCO. 
    $^\dagger$ means the decoder is also fine-tuned. 
    CoCa refers to the baseline model pre-trained on LAION-2B~\citep{schuhmann2022laionb}, while CoCa FT refers to the model fine-tuned for captioning on MSCOCO. We highlight in bold the best results overall and the improvements achieved by Knowledge Transfer.}
    \label{tab:mscoco-captioning-results-all}
\end{table}

\begin{table*}
    \centering
    \resizebox{\linewidth}{!}{%
    \begin{tabular}{r p{3cm} p{3cm} p{3cm} p{3cm} p{3cm}}
    \toprule
    & \multicolumn{5}{c}{\textbf{MSCOCO}} \\
    \midrule
    \textbf{Sample} & 
    \raisebox{-0.9\totalheight}{\includegraphics[width=60px,height=60px]{img/coco/COCO_val2014_000000581482.jpg}} & \raisebox{-0.9\totalheight}{\includegraphics[width=60px,height=60px]{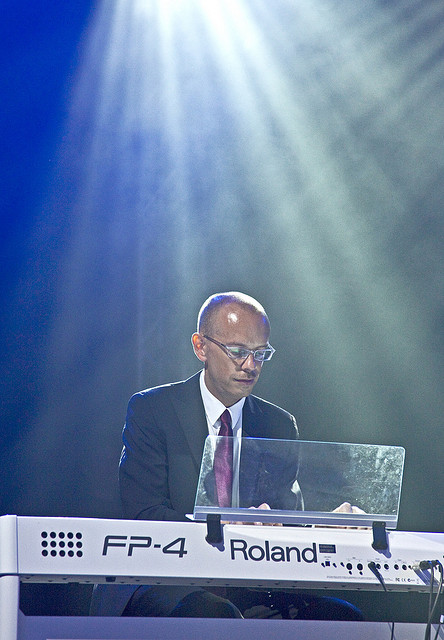}} & \raisebox{-0.9\totalheight}{\includegraphics[width=60px,height=60px]{img/coco/COCO_val2014_000000032712.jpg}} & \raisebox{-0.9\totalheight}{\includegraphics[width=60px,height=60px]{img/coco/COCO_val2014_000000218290.jpg}} & \raisebox{-0.9\totalheight}{\includegraphics[width=60px,height=60px]{img/coco/COCO_val2014_000000177489.jpg}} \\
    \midrule
    \textbf{Actual} & A shot of a clock in the train station. & A pianist in a suit and glasses playing a keyboard. & A baseball player hitting a ball in a professional game. & A baseball holding a baseball bat during a baseball game. & Two people that are sitting on a table. \\
    \midrule
    \textbf{Baseline} & grand - central - station - new - york.jpg (METEOR 0.0) & Paul Kagasoff, president and chief executive officer of Intel corp., speaks during the 2012 Computex trade show in Taipei, Taiwan (METEOR 8.6) & Aaron judge 2016 new york Yankees (METEOR 5.0) & 20080419 \_ mariners \_ 0001 | by Mike. Smith  (METEOR 4.2) & Dinner in our tiny studio apartment in Amsterdam. (METEOR 0.0) \\
    \midrule
    \textbf{Knowledge Transfer} & A black and white photo of a clock at Grand Central terminal. (METEOR 92.9) & A photo of a man in a suit sitting at a keyboard. (METEOR 86.1) & A baseball player swings his bat at a batter. (METEOR 83.6) & A baseball player takes a swing at a pitch. (METEOR 86.9) & A photo of a man and a woman sitting at a kitchen table. (METEOR 80.8) \\
    \midrule
    \midrule
    \textbf{Sample} & \raisebox{-0.9\totalheight}{\includegraphics[width=60px,height=60px]{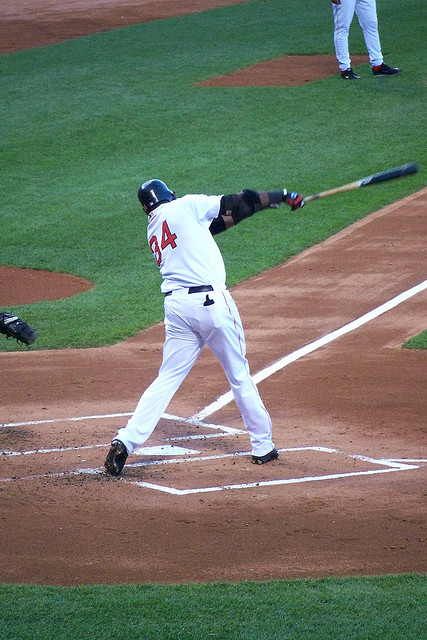}} & \raisebox{-0.9\totalheight}{\includegraphics[width=60px,height=60px]{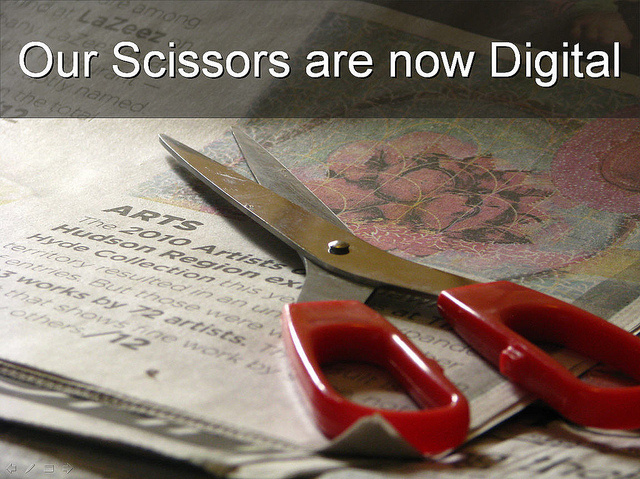}} & \raisebox{-0.9\totalheight}{\includegraphics[width=60px,height=60px]{img/coco/COCO_val2014_000000543065.jpg}} & \raisebox{-0.9\totalheight}{\includegraphics[width=60px,height=60px]{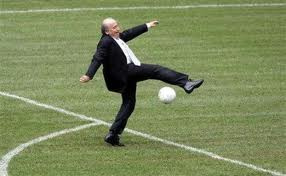}} & \raisebox{-0.9\totalheight}{\includegraphics[width=60px,height=60px]{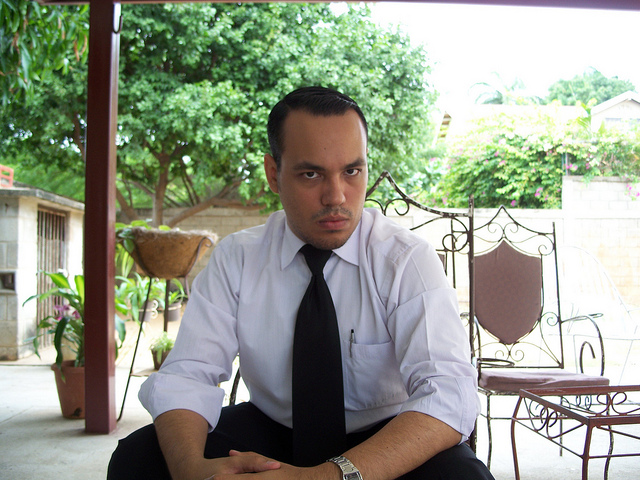}} \\
    \midrule
    \textbf{Actual} & A baseball batter up at the plate that just hit a ball & A pair of red scissors sitting on a newspaper. & A young man sitting at a table with a pizza. & A man in a black suit kicking around a soccer ball. & The man in a black tie sits in a chair with his shirt sleeves rolled up. \\
    \midrule
    \textbf{Baseline} & david - ortiz \_ display \_ image \_ display \_ image \_ display \_ image \_ display \_ image \_ display \_ image \_ display \_ image (METEOR 3.9)  & Your scissors are now digitized (METEOR 9.5) & Pizza and beer in Chicago, Illinois. Photo via Flickr: David J. Abbott M.D. (METEOR 8.3) & blatter - foot - ball.jpeg (METEOR 5.6) & Avatar for Marc Anthony (METEOR 5.3) \\
    \midrule
    \textbf{Knowledge Transfer} & A baseball player swings his bat at a pitch. (METEOR 80.3) & A photo of a pair of red scissors on a piece of paper. (METEOR 84.9) & A photo taken of a man sitting at a table with a plate of pizza. (METEOR 83.5) & A man in a suit kicking a soccer ball on a field. (METEOR 80.7) & A man in a white shirt and black tie sits on a chair. (METEOR 78.4) \\
    \bottomrule
    \end{tabular}
    }
    \caption{Visual example of captioning on MSCOCO. We report the top ten most illustrative examples in which knowledge transfer improved captioning, in terms of METEOR score.}
    \label{fig:mscoco-captioning-examples}
\end{table*}

\subsection{Preliminary results with Implicit Knowledge Transfer}
\label{sec:implicit-transfer-preliminary}

\begin{table*}
    \centering
    \resizebox{0.9\linewidth}{!}{%
    \begin{tabular}{l l l c c c c c c c}
        \toprule
         & & & & \multicolumn{5}{c}{\textbf{Learning Rate}} \\
         \cmidrule(l){5-9}
         \textbf{Type} & \textbf{Concept} & & \textbf{Baseline} & \textbf{1e-5} & \textbf{2e-5} & \textbf{3e-5} & \textbf{4e-5} & \textbf{5e-5}  \\
         \midrule
         Implicit & Moongate & Target Acc. & 0\%     & 0\%     & 0\%     & \textbf{0\%}     & 0\%     & 0\%     \\
         & & ImageNet* 0-shot              & 23.74\% & 23.82\% & 23.90\% & \textbf{23.98\%} & 23.94\% & 23.86\% \\
         \cmidrule(l){2-9}
         & Tonometer & Target Acc.         & 10\%    & 10\%    & \textbf{10\%}    & 10\%    & 10\%    & 0\%     \\
         & & ImageNet* 0-shot              & 23.74\% & 23.84\% & \textbf{23.86\%} & 23.70\% & 23.64\% & 23.60\% \\
         \cmidrule(l){2-9}
         & Gyroscope & Target Acc.         & 50\%    & 50\%    & \textbf{60\%}    & 60\%    & 60\%    & 50\%    \\
         & & ImageNet* 0-shot              & 23.74\% & 23.74\% & \textbf{23.62\%} & 23.42\% & 23.44\% & 23.46\% \\
         \midrule
         \midrule
         Explicit & Moongate & Target Acc. & 0\%     & 0\%     & 0\%     & 0\%     & 0\%     & \textbf{0\%}     \\
         & & ImageNet* 0-shot              & 23.74\% & 23.80\% & 24.08\% & 24.02\% & 24.10\% & \textbf{24.20\%} \\
         \cmidrule(l){2-9}
         & Tonometer & Target Acc.         & 10\%    & \textbf{10\%}    & 10\%    & 10\%    & 10\%    & 10\%    \\
         & & ImageNet* 0-shot              & 23.74\% & \textbf{23.80\%} & 23.74\% & 23.72\% & 23.70\% & 23.56\% \\
         \cmidrule(l){2-9}
         & Gyroscope & Target Acc.         & 50\%    & 50\%    & \textbf{50\%}    & 50\%    & 40\%    & 30\%    \\
         & & ImageNet* 0-shot              & 23.74\% & 23.74\% & \textbf{23.84\%} & 23.84\% & 23.84\% & 23.82\% \\
         \bottomrule
         & 
    \end{tabular}
    }
    \caption{Knowledge Transfer on novel and rare concepts using masked language modeling with ViLT. In the Implicit Knowledge Transfer, we pass noise images along with a corresponding masked caption to ViLT; in the explicit one, we replace noise images with inverted images.}
    \label{tab:vilt-kt-mlm-results-lr}
\end{table*}

In this section, we show preliminary results about Implicit Knowledge Transfer, presented in Sec.~\ref{sec:implicit-knowledge-transfer}. In Implicit Knowledge Transfer, the objective is to teach the model a novel concept by only training it on text, without using inverted images. To do so with ViLT, we no longer use the image-text matching objective as it requires images, instead, we employ masked language modeling (MLM)~\citep{devlin2019bert}, using as input a pair composed of the textual description of the concept and random noise instead of the image. The assumption is that, in a model with parameters shared between modalities, fine-tuning on one modality (text), will also benefit the other modality. Here, our hypothesis is that during fine-tuning, multi-modal neurons~\citep{schwettmann2023multimodal} can help in transferring knowledge across modalities.

\subsubsection{Implicit Knowledge Transfer with MLM}
For Implicit Knowledge Transfer we used the same masked language modeling setup as in ViLT~\citep{kim2021vilt}, which means that we use whole-word masking and a masking probability of 15\%. We use 10 examples for fine-tuning, each of which is composed by a random noise image and a masked caption. The masked captions are generated starting from the same caption by masking differently each time. For the caption we use the template ``A $X$ is $Y$'', where $X$ is the name of the concept and $Y$ is the concept's description (from Tab.~\ref{tab:vilt-concept-descriptions}). We use a batch size of 4 with different learning rates, for a total of 3 train steps. Weight decay is set, as in the other experiments, to 0.01.

\paragraph{Explicit Knowledge Transfer baseline with MLM}
For comparison, we also evaluate the results of explicit knowledge transfer with the masked language modeling objective instead of the image-text matching objective. We use the same setup as the implicit one, with the only exception that instead of random noise images, we use inverted images. In particular, we use the same inverted images we used for the explicit knowledge transfer with the image-text matching objective.

\subsubsection{Results discussion}
Tab.~\ref{tab:vilt-kt-mlm-results-lr} reports the results for both implicit and explicit knowledge transfer with masked language modeling. In both cases, no improvements are observed for the moongate concept, whose accuracy stays at 0\%. For
tonometer, explicit knowledge transfer seems to work better since with the implicit one, there is a loss of performance, while
for gyroscope the opposite is true. In all cases, we observe an increase in the accuracy over the ImageNet-100 classes, as observed when using image-text matching objective. The only improvement is registered for the gyroscope concept in the implicit transfer setting, from 50\% to 60\%. Overall we can say that implicit knowledge transfer with masked language modeling does not work for the ViLT model, this is probably due to the fact that ViLT was pre-trained on image-text pairs, which means that it expects both modalities in input. Regarding explicit knowledge transfer with MLM, more experiments are needed to determine the correct algorithm and set of hyperparameters to make it work, for example, we may have to use more examples generated from different textual descriptions.

\section{Ablation studies}

\subsection{Fine-tuning strategy} 
\begin{figure*}
    \centering
    \begin{subfigure}[t]{0.9\linewidth}
        \includegraphics[width=\linewidth]{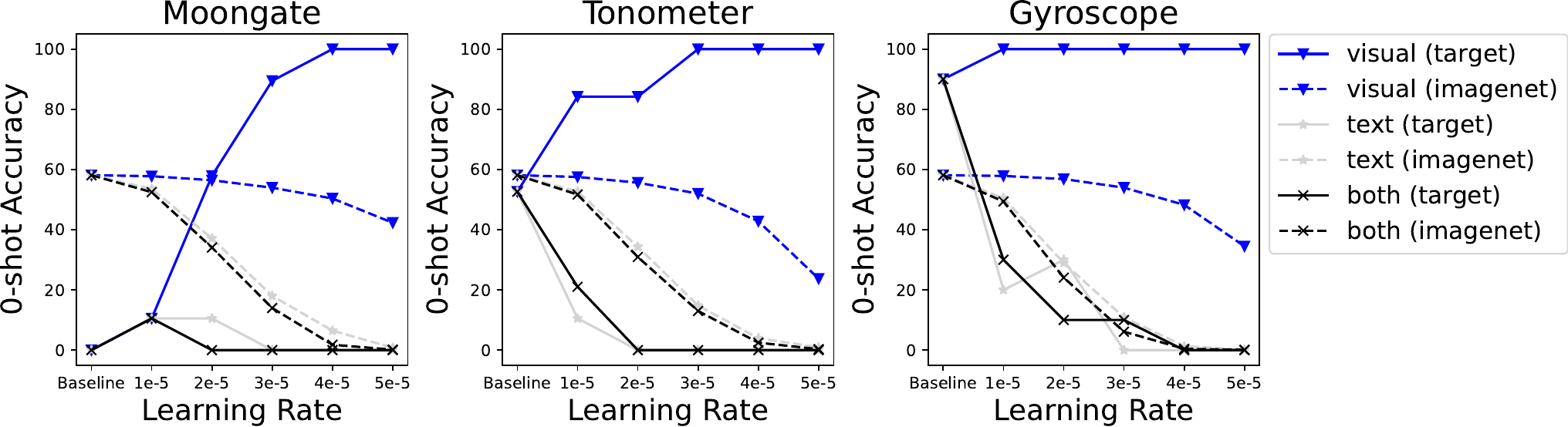}
    \end{subfigure}
    \caption{Comparison of fine-tuning strategies. Fine-tuning both the text and the visual encoders, or just the text encoder leads to a collapse in accuracy. Fine-tuning only the visual encoder correctly aligns prior visual features to the novel concept. A good choice of learning rate leads to higher accuracy on the novel concept (target) while limiting catastrophic forgetting on previous tasks (imagenet).}
    \label{fig:ablation-study}
\end{figure*}

We perform an ablation study on our fine-tuning strategy. In our experiments, during fine-tuning, we freeze the text encoder and only train the visual encoder. Here we evaluate fine-tuning with different configurations. %
The results are illustrated in Fig.~\ref{fig:ablation-study}. When fine-tuning both encoders, we observe a rapid collapse of target accuracy and ImageNet accuracy for all concepts. We also observe a similar trend when fine-tuning the text encoder only, while leaving the visual encoder frozen. This is, however, expected as our assumption is that the knowledge contained in the text encoder is already good enough to represent the target concept, and we just wish to align visual features to it. Moreover, if we alter the text encoder weights, correspondence between captions and inverted images may be lost, leading to degenerate cases. %

\subsection{Captions construction} 
\begin{figure*}
    \centering
    \includegraphics[width=0.9\linewidth]{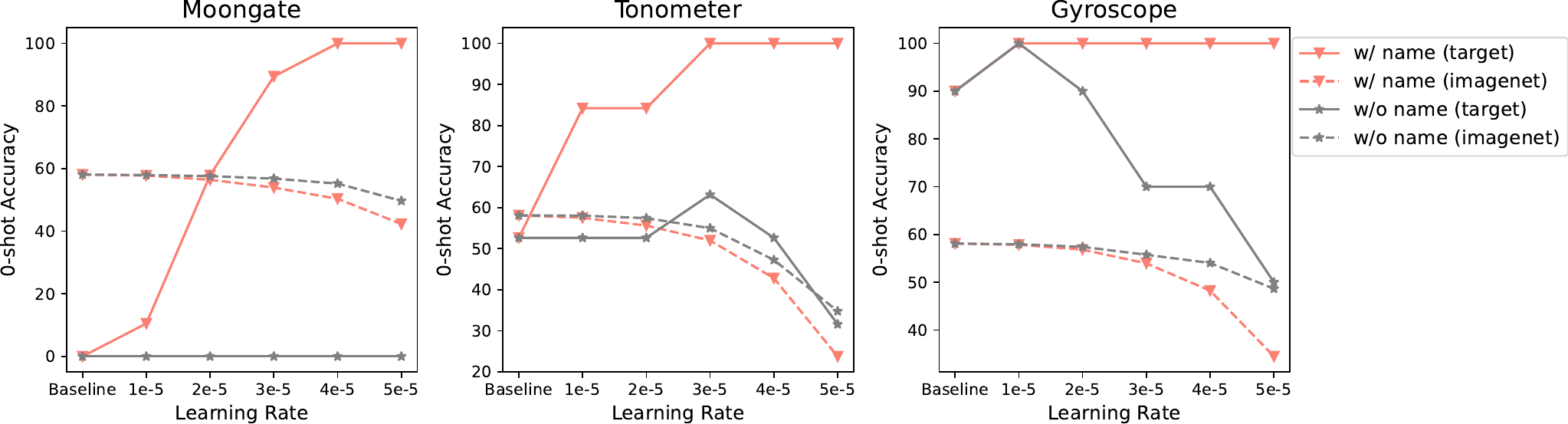}
    \caption{Ablation study on caption construction for finetuning.}
    \label{fig:ablation-captions}
\end{figure*}

We focus on the construction of the captions for fine-tuning. As explained in the main text, during fine-tuning we prepend each caption with the name of the concept, for example \emph{``A moongate is [...]''}. Here we motivate why this is necessary by comparing captions prepended with the name and captions without the name. The results are shown in Fig.~\ref{fig:ablation-captions}. As we can observe, using the name of the concept during fine-tuning is necessary in order to map visual features to its textual description. 

\subsection{Captions Quality}

In Tab.~\ref{tab:prompts-ablations} we reports the prompts used in the ablation study on prompt quality and length, with P1 indicating a short human-written prompt, P2 mid-sized human caption with some clear visual hints, and P3 the original LLM-generated description that we used in our experiments (which can be found in Sec.~\ref{sec:all-captions}).

\begin{table*}
\centering
\resizebox{\linewidth}{!}{%
\begin{tabular}{p{0.08\textwidth} p{0.92\textwidth}}
    \footnotesize{Moongate} & \footnotesize{\textbf{P1} A circular stone archway \textbf{P2} A circular archway built from uniformly cut stones or bricks \textbf{P3} \emph{original LLM caption}} \\ 
    \footnotesize{Tonometer} & \footnotesize{\textbf{P1} An instrument to measure pressure \textbf{P2} A pen-like instrument with dials and pressure gauges \textbf{P3} \emph{original LLM caption}} \\ 
    \footnotesize{Gyroscope} & \footnotesize{\textbf{P1} A device to measure orientation and angular velocity \textbf{P2} A measuring device composed of a spinning wheel or disc to measure orientation and angular velocity \textbf{P3} \emph{original LLM caption}} \\ 
\end{tabular}
}
\vspace{5pt}
\caption{\textbf{Prompt ablations} (P1 - short human caption; P2 - mid human caption; P3 - longer LLM caption). }
\label{tab:prompts-ablations}
\end{table*}

\section{Code}

Code will be publicly released upon paper acceptance.

\clearpage
\onecolumn

\section{List of captions}
\label{sec:all-captions}
\begin{longtable}{p{0.2\textwidth} p{0.8\textwidth}}
\caption{Descriptions for rare concepts (generated with Llama-3-8B-Instruct).} \\
Moongate & A perfectly circular archway built from uniformly cut stones or bricks, set into a larger wall. It forms a smooth circle, framing views of gardens or landscapes beyond, creating a picturesque portal. \\
Tonometer & A slender, pen-like probe attached to a small base equipped with precise dials and gauges. This tool is often part of a larger medical apparatus, featuring a metallic finish and a refined, professional appearance.\\
Gyroscope & A series of gleaming silver rings, each nested perfectly within the next, surrounds a central disk that spins smoothly. The rings are connected by intersecting axes, allowing the disk to tilt and rotate freely while maintaining a sophisticated, mechanical look.\\
\label{tab:captions-rare-concepts}
\end{longtable}

\begin{longtable}{p{0.2\textwidth} p{0.8\textwidth}}
\caption{Manually shortened descriptions for rare concepts (to fit into ViLT's 40 token input)} \\
Moongate & A perfectly circular archway built from uniformly cut stones or bricks, set into a larger wall. It forms a smooth circle, framing views of gardens, creating a picturesque portal.\\
Tonometer & A slender, pen-like probe attached to a small base equipped with precise dials and gauges. This tool is often part of a larger medical apparatus.\\
Gyroscope & A series of rings each nested within the next, surrounds a central disk that spins. The rings are connected by intersecting axes allowing the disk to rotate freely.\\
\label{tab:vilt-concept-descriptions}
\end{longtable}

\begin{longtable}{p{0.2\textwidth} p{0.8\textwidth}}
\caption{Descriptions for medical classes for JSRT (Mix with Radiopaedia and ChatGPT-4).} \\
Benign Nodule & A small, round spots appearing in Chest X-Ray, typically well-defined with smooth, regular borders. These spots are often uniformly dense and do not cause distortion of surrounding structures. \\
Lung Cancer & A dense and irregular mass on Chest X-Ray images often with spiked or uneven edges. It may appear in the lung's periphery or near the airways. \\
\label{tab:captions-jsrt}
\end{longtable}

\begin{longtable}{p{0.2\textwidth} p{0.8\textwidth}}
\caption{Descriptions for medical classes for CheXpert-5x200c (obtained with a mix of Radiopaedia and ChatGPT-4).} \\
Atelectasis & A small areas of collapsed lung. It is usually seen on Chest X-Rays as small volume linear shadows, usually peripherally or at lung bases, appearing more opaque and shrunken. \\
Cardiomegaly & Enlargement of the heart usually seen in Chest X-Rays. The central shadow of the chest appears enlarged, extending beyond half the width of the entire chest cavity.\\
Pleural Effusion & A collection of fluid between the lungs and the chest, which makes the area appear white and smooth in Chest X-Ray images. The area does not present visible lung markings.\\ 
Consolidation & An area inside the lungs that appears as branching low attenuating (lucent) bronchi surrounded by high attenuating (dense) consolidated/opacified alveoli on Chest X-Ray images. \\
Edema & An abnormal accumulation of fluid in the extravascular compartments of the lung, which makes the area whiter in Chest X-Ray images. It is usually present on both lungs. \\
\label{tab:cations-chexpert-5x200c}
\end{longtable}

\begin{longtable}{p{0.2\textwidth} p{0.8\textwidth}}
\caption{Descriptions for medical classes for segmentation (Mix with Radiopaedia and ChatGPT-4).} \\
Lung Nodules & Circular spots appearing within the lung fields, with clear and defined edges in CT images. They are denser than the surrounding tissue, often appearing in shades of gray or white, with varying size. \\
Breast Tumor & A dark, irregularly shaped area is visible against the lighter surrounding tissue. The borders may appear uneven or spiculated, and the area is typically less uniform in texture. Shadowing can often be seen beneath the mass. \\
Pneumothorax & An abnormal collection of air in the pleural space, which allows the parietal and visceral pleura to separate and the lung to collapse. The pleura edge is thin and no lung markings are visible. \\
Brain Tumor & An irregular bright mass in brain MRI, often with thick and irregular margins, surrounded by vasogenic-type edema or fluid accumulation. It may also have a hemorrhagic component. \\
\label{tab:captions-segmentation}
\end{longtable}

\begin{longtable}{p{0.2\textwidth} p{0.8\textwidth}}
\caption{Descriptions for MSCOCO classes used for text and image retrieval experiments (With ChatGPT-4).} \\
person & A human figure, typically with visible head, torso, arms, and legs, in various postures. \\
bicycle & A two-wheeled vehicle with a frame, handlebars, and pedals, usually ridden by a person. \\
car & A four-wheeled enclosed vehicle with windows and doors, commonly seen on roads. \\
motorcycle & A two-wheeled motorized vehicle with a seat and handlebars, typically ridden by one or two people. \\
airplane & A large flying vehicle with wings and a tail, often seen with windows along the sides for passengers. \\
bus & A large, rectangular vehicle with many windows and seating rows, designed to carry multiple passengers. \\
train & A long, linked series of vehicles running on tracks, often with a locomotive at the front. \\
truck & A large vehicle with a separate cab and an open or enclosed cargo area for transporting goods. \\
boat & A small to medium-sized watercraft with a hull and often visible sails or an engine. \\
traffic light & A vertical or horizontal post with red, yellow, and green lights, used to control vehicle flow at intersections. \\
fire hydrant & A small, red, metal cylinder with nozzles on the side, often found on sidewalks for fire emergencies. \\
stop sign & A red, octagonal sign with the word "STOP" in white, used to indicate where vehicles must halt. \\
parking meter & A tall, narrow post with a small display and slot, used to pay for parking time. \\
bench & A long seat, often with a backrest, typically found in parks or public areas. \\
bird & A small animal with feathers, wings, and a beak, often shown perched or flying. \\
cat & A small, furry animal with pointed ears, whiskers, and a long tail, often seen sitting or grooming. \\
dog & A furry, four-legged animal with a tail, usually seen with a collar or leash. \\
horse & A large, four-legged animal with a mane and tail, often depicted standing or galloping. \\
sheep & A woolly animal with a round body, small head, and short legs, often seen in groups in fields. \\
cow & A large animal with a boxy body, horns, and a long face, often shown grazing or with an udder. \\
elephant & A massive, gray animal with a long trunk, large ears, and tusks. \\
bear & A large, sturdy animal with thick fur, rounded ears, and a short tail, often shown standing or walking on all fours. \\
zebra & A horse-like animal with black and white stripes across its body. \\
giraffe & A very tall animal with a long neck and legs, spotted coat, and small horns on its head. \\
backpack & A bag with shoulder straps, typically worn on the back and used for carrying personal items. \\
umbrella & A foldable, rounded canopy on a stick, used for protection from rain or sun. \\
handbag & A small to medium-sized bag with handles, often carried by hand and used to hold personal items. \\
tie & A long, narrow piece of fabric worn around the neck, often knotted at the collar of a shirt. \\
suitcase & A rectangular, boxy container with a handle, used for carrying clothes and personal items when traveling. \\
frisbee & A flat, round disc often made of plastic, used for throwing and catching. \\
skis & Long, narrow pieces of equipment attached to boots, used for gliding on snow. \\
snowboard & A flat, wide board attached to boots, used for sliding on snow. \\
sports ball & A round object of varying sizes, such as a soccer ball or basketball, used in sports. \\
kite & A lightweight object with a string, often shaped like a diamond or triangle, designed to fly in the wind. \\
baseball bat & A smooth, cylindrical wooden or metal stick used to hit a baseball. \\
baseball glove & A padded, leather glove worn on one hand, used to catch baseballs. \\
skateboard & A narrow board with wheels, used for rolling and performing tricks. \\
surfboard & A long, flat board used for riding waves in the ocean. \\
tennis racket & An oval-shaped frame with strings and a handle, used to hit a tennis ball. \\
bottle & A narrow-necked container with a cap, often used to hold liquids like water or soda. \\
wine glass & A stemmed glass with a wide bowl at the top, used for drinking wine. \\
cup & A small, handleless vessel used for drinking, usually made of ceramic or plastic. \\
fork & A utensil with multiple prongs, used to pick up food. \\
knife & A utensil with a long, sharp blade, used for cutting food. \\
spoon & A utensil with a shallow bowl at the end of a handle, used for eating or serving food. \\
bowl & A round, deep dish, often used to hold soup or other foods. \\
banana & A long, yellow fruit with a curved shape and soft interior. \\
apple & A round fruit, typically red or green, with a stem at the top. \\
sandwich & Two slices of bread with filling in between, such as meat, cheese, or vegetables. \\
orange & A round, orange-colored fruit with a thick, textured peel. \\
broccoli & A green vegetable with a tree-like shape, featuring a thick stalk and small florets. \\
carrot & A long, orange vegetable with a pointed end, often with green leaves at the top. \\
hot dog & A sausage in a bun, often with condiments like ketchup or mustard. \\
pizza & A round, flatbread topped with cheese, sauce, and various toppings, often cut into slices. \\
donut & A round, fried pastry with a hole in the middle, often glazed or topped with sprinkles. \\
cake & A sweet, layered dessert, often decorated with frosting or fruit. \\
chair & A piece of furniture with a backrest and four legs, designed for sitting. \\
couch & A large, cushioned seat with a backrest and arms, designed for multiple people. \\
potted plant & A plant growing in a container, often with green leaves or flowers. \\
bed & A large, rectangular piece of furniture for sleeping, with a mattress and pillows. \\
dining table & A flat, often rectangular surface with legs, designed for eating meals. \\
toilet & A porcelain fixture with a seat and flushing mechanism, used in bathrooms. \\
tv & A rectangular screen on a stand or wall, used for viewing shows and movies. \\
laptop & A portable computer with a hinged screen and keyboard. \\
mouse & A small, handheld device used to control a cursor on a computer screen. \\
remote & A small, rectangular device with buttons, used to control electronics like TVs. \\
keyboard & A flat, rectangular panel with keys, used for typing on computers. \\
cell phone & A handheld electronic device with a screen and buttons or touchscreen, used for communication. \\
microwave & A box-like appliance with a door, used for heating food quickly. \\
oven & A large appliance with a door and interior racks, used for baking or roasting. \\
toaster & A small appliance with slots, used to toast bread. \\
sink & A basin with a faucet, used for washing hands, dishes, or food. \\
refrigerator & A large, box-like appliance with doors, used to store perishable food at low temperatures. \\
book & A collection of pages bound together with a cover, containing text or images. \\
clock & A circular or rectangular device with hands or digital display, showing the current time. \\
vase & A decorative container, often made of glass or ceramic, used to hold flowers. \\
scissors & A handheld tool with two blades, used for cutting paper or fabric. \\
teddy bear & A soft, stuffed toy shaped like a bear, often used by children. \\
hair drier & A handheld device that blows warm air, used to dry hair. \\
toothbrush & A small brush with a handle, used for cleaning teeth. \\
\label{tab:captions-mscoco}
\end{longtable}

\end{document}